\documentclass{article}

\usepackage{arxiv}

\usepackage{amsmath,amssymb}
\usepackage[utf8]{inputenc} 
\usepackage[T1]{fontenc}    
\usepackage{hyperref}       
\usepackage{url}            
\usepackage{booktabs}       
\usepackage{amsfonts}       
\usepackage{nicefrac}       
\usepackage{microtype}      
\usepackage{lipsum}
\usepackage{graphicx}
\usepackage[ruled,vlined,linesnumbered]{algorithm2e}
\usepackage{tabularx}
\usepackage{multirow}
\usepackage{threeparttable}
\usepackage{subfigure}
\usepackage{subcaption}
\usepackage{xurl}
\usepackage{enumitem}
\graphicspath{ {./figures/} }

\title{ACD-U: Asymmetric co-teaching with machine unlearning for robust learning with noisy labels}

\author{
Reo Fukunaga \hspace{.2cm}
Soh Yoshida\thanks{Corresponding author} \hspace{.2cm}
Mitsuji Muneyasu \\
Kansai University \\
3-3-35 Yamate-cho, Suita, Osaka 564-8680, Japan \\
\texttt{sohy@kansai-u.ac.jp}
}

\begin{document}
\maketitle
\begin{abstract}
Deep neural networks are prone to memorizing incorrect labels during training, which degrades their generalizability. Although recent methods have combined sample selection with semi-supervised learning (SSL) to exploit the memorization effect---where networks learn from clean data before noisy data---they cannot correct selection errors once a sample is misclassified. To overcome this, we propose asymmetric co-teaching with different architectures (ACD)-U, an asymmetric co-teaching framework that uses different model architectures and incorporates machine unlearning. ACD-U addresses this limitation through two core mechanisms. First, its asymmetric co-teaching pairs a contrastive language-image pretraining (CLIP)-pretrained vision Transformer with a convolutional neural network (CNN), leveraging their complementary learning behaviors: the pretrained model provides stable predictions, whereas the CNN adapts throughout training. This asymmetry, where the vision Transformer is trained only on clean samples and the CNN is trained through SSL, effectively  mitigates confirmation bias. Second, selective unlearning enables post-hoc error correction by identifying incorrectly memorized samples through loss trajectory analysis and CLIP consistency checks, and then removing their influence via Kullback--Leibler divergence-based forgetting. This approach shifts the learning paradigm from passive error avoidance to active error correction. Experiments on synthetic and real-world noisy datasets, including CIFAR-10/100, CIFAR-N, WebVision, Clothing1M, and Red Mini-ImageNet, demonstrate state-of-the-art performance, particularly in high-noise regimes and under instance-dependent noise. The code is publicly available at \url{https://github.com/meruemon/ACD-U}. 
\end{abstract}


\section{Introduction}
Deep neural networks (DNNs) exhibit high representational capacity and have achieved considerable success in various applications, including image classification. However, fully leveraging their potential requires large-scale datasets with accurate labels. Common methods for constructing such datasets, such as crowdsourcing~\cite{welinder_2010_neural} and web-based data collection using search engines~\cite{chen_2013_IEEE, schroff_2010_IEEE, niu_2015_IEEE}, inevitably introduce label noise. Reported noise rates for these datasets range from 8 to 35.8\%~\cite{xiao_2015_clothing, song_2019_PMLR, li_2017_webvision, Song_2020_prestopping}. DNNs tend to overfit to noisy labels~\cite{Zhang_2021_understanding,Devansh_2017_PMLR}, causing them to learn incorrect patterns and significantly degrading their generalization performance on unseen data.

To address this challenge, researchers have developed various approaches for learning with noisy labels (LNL) that leverage the learning characteristics of DNNs~\cite{Liu_2020_PMLR, Liu_2022_PMLR, Patrini_2017_IEEE, Han_2018_Co-teaching}. A key discovery in this field is the memorization effect (ME)~\cite{Devansh_2017_PMLR}, which describes how DNNs preferentially learn from clean samples in the early training stages before eventually memorizing noisy samples as training progresses. This property enables loss-based sample selection, as clean samples tend to exhibit lower loss values than noisy ones.

Co-teaching~\cite{Han_2018_Co-teaching} is a prominent method that leverages this discovery. In this approach, two independent networks act as mutual teachers; each network selects low-loss samples to train its peer, thereby mitigating the confirmation bias~\cite{yu_2019_PMLR}---the tendency to reinforce incorrect predictions---inherent in single-model learning.

Building on this success, DivideMix~\cite{Li_2020_DivideMix} further improved performance by integrating sample selection with semi-supervised learning (SSL), establishing what is now a mainstream paradigm~\cite{cordeiro_2023_LongReMix, liu_2025_PLReMix}. DivideMix and its extensions~\cite{cordeiro_2023_LongReMix, liu_2025_PLReMix, Zhang_2023_Rankmatch, xiao_2022_promix, Liang_2024_SV-Learner, Feng_2024_NoiseBox, Feng_2024_Clipcleaner, kim_2024_LSL, li_2025_RML} classify samples as clean or noisy based on their loss values, using the reliable samples as labeled data and treating the unreliable ones as unlabeled data.

However, these SSL-based methods face several challenges. First, because the labels of samples selected as “clean” are used directly, it is difficult to correct those that were initially misidentified. While recent methods such as ProMix~\cite{xiao_2022_promix} and RankMatch~\cite{Zhang_2023_Rankmatch} aim to enhance selection accuracy through progressive confidence enhancement or multi-model consensus, recovering from initial errors remains a significant challenge. Second, incorrect sample selections made during early training, when model predictions are unstable, can propagate and negatively impact the entire learning process~\cite{takeda_2021_IEEE}. Consequently, models continue to memorize noisy labels and accumulate errors over time.

Recently, large-scale pretrained models, such as those trained on ImageNet, or vision-language models (VLMs), such as CLIP, have gained attention as a partial solution~\cite{ahn_2023_arXiv, wang_2025_arXiv, Feng_2024_Clipcleaner}. However, existing methods typically use these models only for initialization~\cite{ahn_2023_arXiv} or as static zero-shot classifiers~\cite{Feng_2024_Clipcleaner}. Therefore, they fail to leverage the different learning behaviors between pretrained models, which exhibit high accuracy early on, and conventional, randomly initialized models.

Meanwhile, in the field of machine unlearning, techniques have been developed to selectively remove the influence of specific data from trained models~\cite{Kurmanji_2023_SCRUB, bourtoule_2021_sisa, izzo_2021_newton}. Although originally designed for privacy protection and compliance with the ``right to be forgotten'' legislation~\cite{mantelero_2013_Elsevier, bill_ab_375}, this technology offers a promising approach for the post-hoc correction of incorrectly memorized information in noisy-label learning.

This paper presents the \textbf{asymmetric co-teaching with different architectures (ACD)-U} framework, an asymmetric co-teaching approach with different architectures that is enhanced by machine unlearning. This method addresses the aforementioned limitations through two novel, integrated mechanisms:
\begin{enumerate}
\item \textbf{Selective unlearning for error correction:} We introduce machine unlearning to the LNL domain, enabling the post-hoc correction of memorized errors. We propose a selective unlearning method that dynamically identifies and forgets incorrectly learned noisy samples by analyzing model-loss trajectories and checking for consistency with contrastive language-image pretraining (CLIP)’s~\cite{Radford_2021_CLIP} zero-shot predictions.
\item \textbf{ACD:} We implement an asymmetric co-teaching strategy that pairs a CLIP-pretrained vision Transformer (ViT)~\cite{Radford_2021_CLIP} with a conventional CNN. By leveraging the ViT’s pretrained knowledge and consideration of different learning characteristics through an asymmetric training scheme, this design improves early-stage sample selection and mitigates confirmation bias.
\end{enumerate}

The main contributions of this study can be summarized as follows:
\begin{itemize}
   \item We present the first application of machine unlearning to noisy-label learning, enabling the selective forgetting of incorrectly memorized samples for post-hoc error correction.
   \item We propose ACD, an asymmetric co-teaching architecture that leverages the distinct learning characteristics of a pretrained ViT and conventional CNN to suppress noise memorization in early training stages.
   \item We demonstrate state-of-the-art performance across synthetic noise benchmarks (CIFAR-10/100) and real-world datasets (CIFAR-N, WebVision, Clothing1M, Red Mini-ImageNet), with substantial improvements in high-noise scenarios.
   \item We provide quantitative evidence that unlearning and ACD play complementary roles, with unlearning improving accuracy at high noise rates and ACD benefiting low-to-moderate noise conditions.
\end{itemize}

The remainder of this paper is organized as follows: Section~\ref{sec:02} discusses related work, Section~\ref{sec:03} details the ACD-U, Section~\ref{sec:04} validates its effectiveness through comparative experiments and ablation analysis, and Section~\ref{sec:05} concludes the paper and presents directions for future research.

\begin{figure*}[t]          
  \centering
  \includegraphics[width=0.99\textwidth, height=0.45\textheight, keepaspectratio]{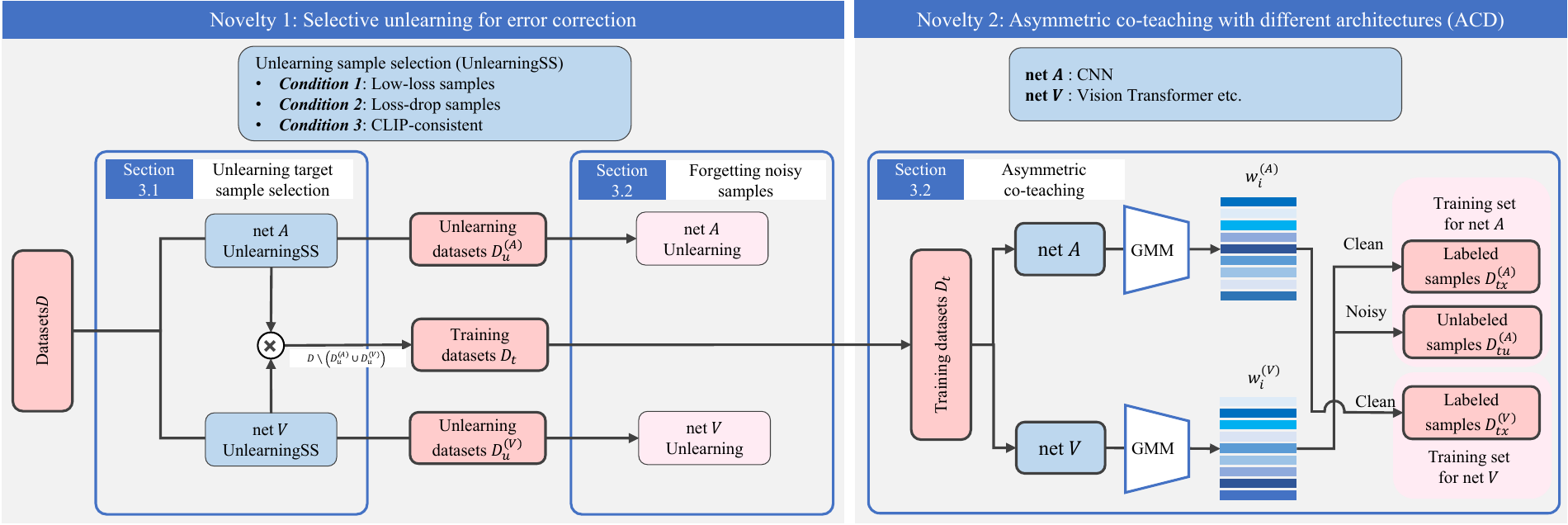}
  \caption{Overview of the ACD-U framework, comprising three main components: (1) unlearning sample selection, (2) sample forgetting via unlearning, and (3) asymmetric co-teaching between different model architectures. Unlearning sample selection is performed every $E_{UP}$  epochs. Subsequently, the sample forgetting process is executed for $E_{UD}$ epochs. The training dataset, $D_t$, used for ACD is constructed by excluding samples from the unlearning target datasets, $D^{(A)}_u$ and $D^{(V)}_u$.}
  \label{fig:overview}
\end{figure*}

\section{Related work}
\label{sec:02}

\subsection{Machine unlearning}
Machine unlearning selectively removes the influence of specific data from trained models. Its fundamental principle is to adjust model parameters to intentionally degrade predictive capability on target data. Early research proposed methods that push prediction probabilities toward a uniform distribution~\cite{golatkar_2020_cvpr} or maximize the output difference from a previously saved model~\cite{guo_2020_ICML}. From a computational efficiency standpoint, the sharded, isolated, sliced, and aggregated model~\cite{bourtoule_2021_sisa} achieves efficient forgetting by partitioning the dataset into multiple shards and independently training models on them. Izzo et al.~\cite{izzo_2021_newton} developed a method for fast forgetting using second-order optimization. More recent practical approaches leverage interactions between multiple models. For instance, knowledge distillation from a bad teacher (Bad-T)~\cite{chundawat_2023_BatT} employs a teacher--student framework with three networks: a competent teacher (trained on clean data), an incompetent teacher (trained on data including the forgetting targets), and a student model. The student learns to mimic the competent teacher's output while diverging from the incompetent teacher's predictions. Similarly, scalable remembering and unlearning unbound (SCRUB)~\cite{Kurmanji_2023_SCRUB} uses contrastive learning to maximize the difference between the prediction distributions of a student and an omniscient teacher for the target samples, while aligning their predictions for retained data to achieve selective forgetting.

However, existing research on machine unlearning has primarily focused on privacy protection and data deletion requests~\cite{cao_2015_towards, ginart_2019_making, guo_2020_certified}, with limited application to improving model learning performance. While some studies have explored theoretical algorithm design~\cite{sekhari_2021_remember} or the verifiability performance~\cite{thudi_2022_necessity}, they have not evaluated how unlearning can enhance a model’s predictive accuracy. Consequently, the application of unlearning to the problem of noisy-label learning remains largely unexplored.

\subsection{Learning with noisy labels}\label{sec:02:02}
The ME, which describes how DNNs learn clean patterns before memorizing noisy labels, is a fundamental principle behind many robust learning approaches. Sample-selection methods leverage the ME to identify clean samples for training. For example, MentorNet~\cite{jiang2018mentornet} introduced a meta-learning framework where a teacher network guides a student’s learning by selecting reliable, low-loss samples. Co-teaching~\cite{Han_2018_Co-teaching} established a dual-network paradigm where two networks mutually teach by selecting small-loss samples for each other, thereby mitigating confirmation bias. Subsequent developments include Co-teaching+~\cite{yu_2019_PMLR}, which incorporates a disagreement mechanism; the overfitting-to-underfitting network (O2U-Net)~\cite{huang2019o2u}, which uses cyclic learning and tracks loss history; and joint training with co-regularization (JoCoR)~\cite{wei2020combating}, which combines co-teaching with co-regularization.

The integration of sample selection with SSL has led to significant progress~\cite{Li_2020_DivideMix, cordeiro_2023_LongReMix, liu_2025_PLReMix, Zhang_2023_Rankmatch, xiao_2022_promix, Liang_2024_SV-Learner, Feng_2024_NoiseBox, Feng_2024_Clipcleaner, kim_2024_LSL, li_2025_RML}.
DivideMix~\cite{Li_2020_DivideMix} pioneered the use of a Gaussian mixture model (GMM) to divide samples into clean and noisy sets, and then applied semi-supervised techniques to effectively utilize both categories. This framework inspired numerous extensions. For instance, ProMix~\cite{xiao_2022_promix} mitigates confirmation bias through progressive confidence-based sample collection, whereas RankMatch~\cite{Zhang_2023_Rankmatch} performs sample selection based on multi-model consensus via confidence voting. These methods, which focus on improving initial sample selection, provide a key point of comparison for our method. Other related methods that follow a similar SSL integration approach include LongReMix~\cite{cordeiro_2023_LongReMix}, which improves confidence through a two-stage learning approach; NoiseBox~\cite{Feng_2024_NoiseBox}, which performs sample selection via subset expansion; and regroup median loss++ (RML++)~\cite{li_2025_RML}, which uses prediction consistency and adaptive thresholds.

Some recent approaches have focused on developing more accurate selection criteria. For example, pseudo-label relaxed contrastive representation learning (PLReMix)~\cite{liu_2025_PLReMix} combines contrastive features and loss values within a 2D GMM, whereas support vector (SV)-learner~\cite{Liang_2024_SV-Learner} integrates support vector machines with contrastive learning. Learning with structural label (LSL)~\cite{kim_2024_LSL} utilizes structured labels and is specifically designed for noisy-label scenarios. Other recent approaches have incorporated pretrained VLMs.
Scaled activation projection (SAP)~\cite{sangamesh2025sap} introduces a form of corrective machine unlearning but applies it only to pretrained models, which differs from our approach of training a CNN from scratch.
CLIPCleaner~\cite{Feng_2024_Clipcleaner} leverages CLIP's zero-shot capability to identify clean samples but treats it as a static external selector rather than a learnable component. By contrast, the denoising fine-tuning framework~\cite{wei_2024_NeurIPS} fully fine-tunes CLIP for multimodal LNL, integrating sample selection and label correction via visual-language alignment. Our use of CLIP differs from that in these methods; we use its ViT as a trainable feature extractor and its text embeddings as a fixed reference, without performing full fine-tuning or prompt learning. Although other studies have explored ViT integration in noisy-label scenarios~\cite{ahn_2023_arXiv, wang_2025_arXiv}, most have used it as a fixed feature extractor rather than an adaptive learning component.

Our work is also distinct from alternative LNL paradigms that do not rely on sample selection. These include methods based on robust loss functions that reduce sensitivity to label noise~\cite{Liu_2020_PMLR, liu_2020_ELR, Liu_2022_PMLR}, noise transition matrix estimation to model label-corruption patterns~\cite{Patrini_2017_IEEE, tanno_2019_IEEE, xia_2019_NIPS}, label correction methods that attempt to fix corrupted labels~\cite{Li_2021_ICCV, Huang_2023_CVPR_TCL, xu_2025_IEEE}, and contrastive learning for robustness enhancement~\cite{Ortego_2021_MOIT, kashiwagi_2024_IEEE}. By designing noise-tolerant loss functions or correcting labels through noise-probability estimation, these approaches differ fundamentally from our sample-selection-based framework.

\section{Proposed method}
\label{sec:03}
We propose ACD-U, a framework for robust learning in noisy-label environments that addresses two key limitations of existing sample-selection methods: (1) the inability to correct memorized labeling errors, and (2) failure to exploit the complementary learning characteristics of pretrained versus randomly initialized networks. The framework combines selective unlearning with asymmetric co-teaching between different architectures. 

As shown in Fig.~\ref{fig:overview}, ACD-U integrates three main components: (1) unlearning target sample selection (Section~\ref{subsec:unlearning_ss}), (2) forgetting of noisy samples (Section~\ref{subsec:selective_unlearning}), and (3) ACD (Section~\ref{subsec:asymmetric_coteaching}). These components operate within a unified training algorithm (Section~\ref{subsec:training_algorithm}).

Let $D=\{(\mathbf{x}_i, y_i)\}_{i=1}^{N}$ be a dataset where $\mathbf{x}_i$ is an input image and $y_i \in \{1, 2, \ldots, C\}$ is a class label that may be corrupted by noise. $N$ is the total number of samples, and $C$ is the number of classes. ACD-U employs two networks: a pretrained ViT (\textbf{net $V$}) and a CNN (\textbf{net $A$}). The parameters of network $l \in \{A, V\}$ are denoted as $\theta_l$.

\subsection{Unlearning target sample selection}\label{subsec:unlearning_ss}
Our method for selecting forgetting targets identifies overfitted noisy samples by integrating three conditions. 

For each network $l \in \{A, V\}$ and sample $(\mathbf{x}_i, y_i)$, we define the cross-entropy loss $\mathcal{L}_s$ as follows:
\begin{equation}\label{eq:cross_entropy}
    \mathcal{L}_s(\mathbf{x}_i; \theta_l) = -\log p_{\theta_l}(y_i|\mathbf{x}_i),
\end{equation}
where $p_{\theta_l}(y_i|\mathbf{x}_i)$ is the probability that network $l$ predicts class $y_i$ for sample $\mathbf{x}_i$.

\subsubsection{\textbf{Condition 1}: Low-loss samples}
Based on the ME, overfitted noisy samples tend to exhibit low loss values. Leveraging this property, we select the following low-loss samples as initial candidates for unlearning:
\begin{equation}\label{eq:low_loss_quantile}
    D_{pl}^{(l)} = \left\{(\mathbf{x}_i,y_i) \in D \mid \mathcal{L}_s(\mathbf{x}_i; \theta_l) < Q_{p_{low}}(\{\mathcal{L}_s(\mathbf{x}_j; \theta_l)\}_{j=1}^N)\right\},
\end{equation}
where $Q_{\alpha}(\cdot)$ is the $\alpha$-quantile function, and $p_{low} \in [0,1]$ is a hyperparameter that controls the proportion of samples selected.

\subsubsection{\textbf{Condition 2}: Loss-drop samples}
Samples that have experienced a drop in loss since the last selection point are likely to have been incorrectly memorized during that interval. We identify these samples using the loss change $\Delta\mathcal{L}_s(\mathbf{x}_i; \theta_l) = \mathcal{L}_s(\mathbf{x}_i; \theta_l) - \mathcal{L}_s^{(k-E_{UP})}(\mathbf{x}_i; \theta_l)$:
\begin{equation}\label{eq:loss_change}
    D_{\Delta l}^{(l)} = \left\{(\mathbf{x}_i,y_i) \in D \mid  \Delta\mathcal{L}_s(\mathbf{x}_i; \theta_l) < Q_{p_{drop}}(\{\Delta\mathcal{L}_s(\mathbf{x}_j; \theta_l)\}_{j=1}^N)\right\},
\end{equation}
where $\mathcal{L}_s^{(k-E_{UP})}(\mathbf{x}_i; \theta_l)$ is the loss from $E_{UP}$ epochs prior to the current epoch $k$, and $p_{drop} \in [0,1]$ controls the proportion of samples selection based on this criterion.

\subsubsection{\textbf{Condition 3}: CLIP-consistent}
The first two conditions may inadvertently select genuinely clean samples. To prevent this, we use a pretrained CLIP model, which is independent of our training process and thus unaffected by the noisy labels, as a filter. We exclude samples whose given label is consistent with CLIP’s zero-shot prediction, as these are likely to be clean:
\begin{equation}\label{eq:CLIP_selection}
   D_{CS} = \left\{(\mathbf{x}_i,y_i) \in D \mid \arg\max p_{CLIP}(\mathbf{x}_i) = y_i\right\},
\end{equation}
where $p_{CLIP}(\mathbf{x}_i)$ is the zero-shot prediction probability from the pretrained CLIP model.  Because this filter relies only on the provided label $y_i$ and CLIP's zero-shot classification result, it is not influenced by models trained on noisy labels, which reduces the likelihood of mistakenly protecting overfitted samples from being forgotten.

\begin{algorithm}[t]
\SetAlgoLined
\caption{Unlearning sample-selection algorithm (UnlearningSS)}
\label{alg:unlearing_smple}

\KwIn{network $l \in \{A, V\}$, network parameters $\theta_l$, dataset $D$, loss before the $E_{UP}$ epoch $\mathcal{L}^{(k-E_{UP})}_s(\mathbf{x}_i; \theta_l)$, $p_{low}$, $p_{drop}$}
\KwOut{unlearning dataset $D_u^{(l)}$}

\ForEach{$(\mathbf{x}_i, y_i) \in D$}{
    $\mathcal{L}_s(\mathbf{x}_i; \theta_l) = -\log p_{\theta_l}(y_i|\mathbf{x}_i)$\;
}

$D_{pl}^{(l)} = \left\{(\mathbf{x}_i, y_i)\in D \;\middle|\; \mathcal{L}_s(\mathbf{x}_i; \theta_l) < \mathrm{Q}_{p_{low}}\left(\{\mathcal{L}_s(\mathbf{x}_j; \theta_l)\}_{j=1}^{N}\right)\right\}$\;

$\Delta \mathcal{L}_s(\mathbf{x}_i; \theta_l) = \mathcal{L}_s(\mathbf{x}_i; \theta_l) - \mathcal{L}^{(k-E_{UP})}_s(\mathbf{x}_i; \theta_l)$\;

$D_{\Delta l}^{(l)} = \left\{(\mathbf{x}_i, y_i)\in D \;\middle|\; \Delta\mathcal{L}_s(\mathbf{x}_i; \theta_l)<\mathrm{Q}_{p_{drop}}\left(\{\Delta\mathcal{L}_s(\mathbf{x}_j; \theta_l)\}_{j=1}^{N}\right)\right\}$\;

$D_{CS} = \left\{(\mathbf{x}_i,y_i)\in D \;\middle|\; \arg\max p_{CLIP}(\mathbf{x}_i)=y_i\right\}$\;

$D_u^{(l)} = (D_{pl}^{(l)}\cup D_{\Delta l}^{(l)}) \setminus D_{CS}$\;

\Return $D_u^{(l)}$\;
\end{algorithm}

\subsubsection{Finalizing the unlearning sets}
By integrating these three conditions, we determine the set of forgetting targets for each network, as outlined in Algorithm~\ref{alg:unlearing_smple}:
\begin{equation}\label{eq:unlearning_selection}
    D_u^{(l)} = \left(D_{pl}^{(l)} \cup D_{\Delta l}^{(l)}\right) \setminus D_{CS}.
\end{equation}
This process yields two sets of samples to be unlearned: $D_u^{(A)}$ for net $A$ and $D_u^{(V)}$ for net $V$.

The training dataset for the ACD component is then constructed by excluding all samples targeted for unlearning by either network:
\begin{equation}\label{eq:train_sample}
   D_t = D\setminus (D_u^{(A)} \cup D_u^{(V)}),
\end{equation}
where $D_t$ is updated every $E_{UP}$ epochs.

Concurrently with selecting the unlearning samples, we save the current network parameters as reference models:
\begin{equation}\label{eq:model_copy}
    \theta_{ref}^{(A)} = \theta_A, \quad \theta_{ref}^{(V)} = \theta_V.
\end{equation}
These reference models store the prediction distributions before the unlearning process begins, enabling effective forgetting by maximizing the difference between the current and reference models' outputs for selected noisy samples. As shown in Eq.~\eqref{eq:model_copy}, these reference models are updated with the current network's parameters each time a new set of forgetting targets is selected.

\subsection{Forgetting noisy samples}\label{subsec:selective_unlearning}
To selectively forget the targeted  noisy samples, we employ a KL-divergence-based loss function inspired by SCRUB~\cite{Kurmanji_2023_SCRUB}. For each network $l \in \{A, V\}$, the unlearning loss for a target sample $\mathbf{x}_u \in D_u^{(l)}$ is defined as
\begin{equation}\label{eq:unlearning}
   \mathcal{L}_{unl}(\mathbf{x}_u; \theta_l, \theta_{ref}^{(l)}) = -T_{unl}^2 \cdot D_{KL}(p_{\theta_{ref}^{(l)}}(\mathbf{x}_u) \| p_{\theta_l}(\mathbf{x}_u)),
\end{equation}
This loss function drives the unlearning process. Here, $p_{\theta_l}(\mathbf{x}_u)$ represents the output probability distribution of the current network ($l$) for a sample targeted for unlearning, whereas $p_{\theta_{ref}^{(l)}}(\mathbf{x}_u)$ is the distribution from the reference model that was saved before the forgetting process began (Eq.~\eqref{eq:model_copy}), The KL divergence ($D_{KL}(\cdot \| \cdot)$) measures the distance between these two distributions, and the hyperparameter $T_{unl} > 0$ controls the intensity of the unlearning.

This function selectively removes the influence of a noisy sample by pushing the current model's predictions  away from the reference model's “pre-forgetting” predictions. The crucial negative sign in the equation achieves this by converting the standard KL divergence minimization into a maximization problem, thereby amplifying the difference between the two distributions. This unlearning process is executed for $E_{UD}$ epochs after each selection of new forgetting targets and runs in parallel with the standard  ACD training on  the filtered dataset $D_t$.

\subsection{Asymmetric co-teaching}\label{subsec:asymmetric_coteaching}
\begin{figure*}[t]
  \centering
  \includegraphics[width=0.99\textwidth, height=0.45\textheight, keepaspectratio]{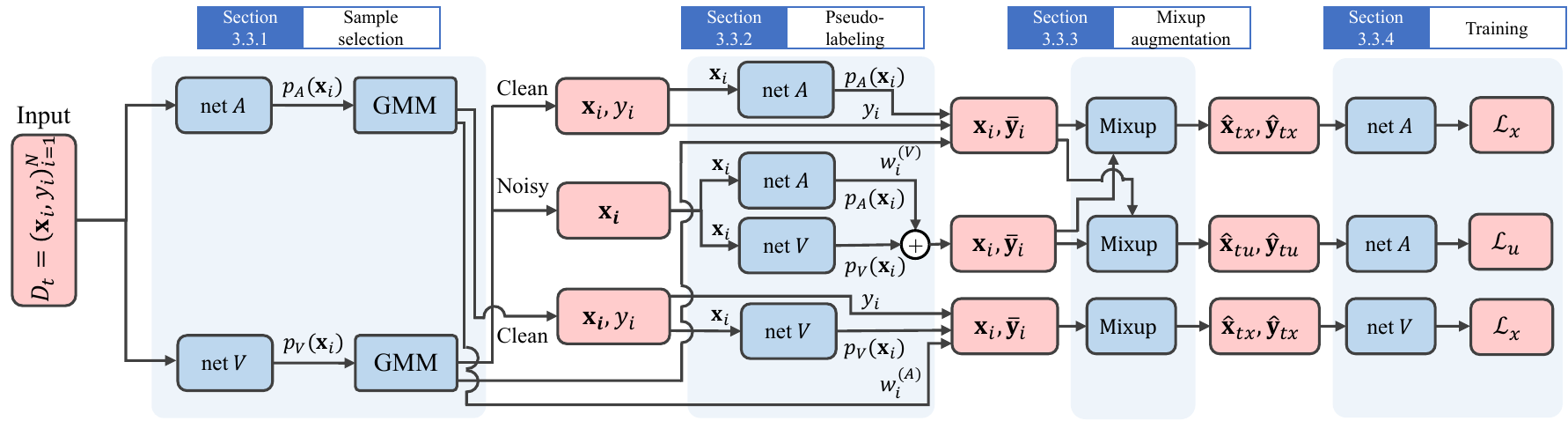}
  \caption{Architecture of asymmetric Co-teaching with different architectures (ACD) framework, which employs an asymmetric training strategy where the pretrained Vision Transformer (net $V$) trains only on labeled samples, whereas the CNN (net $A$) utilizes both labeled and unlabeled samples in a semi-supervised learning pipeline.}
  \label{fig:architecture}
\end{figure*}

Unlike conventional co-teaching methods that use symmetric training for two identical networks~\cite{Li_2020_DivideMix}, we propose an asymmetric approach that leverages the different characteristics of our two models: the pretrained ViT (net $V$) and the randomly initialized CNN (net $A$). This strategy, depicted in Fig.~\ref{fig:architecture}, comprises four phases: sample selection, pseudo-labeling, Mixup, and training.

\subsubsection{\textbf{Phase 1}: Sample selection}
In this phase, the training samples from the dataset $D_t$ are divided into labeled and unlabeled sets for each network. First, we calculate the cross-entropy loss for every sample using Eq.~\eqref{eq:cross_entropy}. We then apply a two-component GMM, following the approach in~\cite{Li_2020_DivideMix}, to the loss distributions of each network.
This process yields the probabilities $w_i^{(A)} \in \mathcal{W}^{(A)}$ and $w_i^{(V)} \in \mathcal{W}^{(V)}$, which represent the likelihood that a given sample is cleanly labeled according to net $A$ and net $V$, respectively. To mitigate confirmation bias, each network uses the probabilities calculated by its peer to select data for its own training. The labeled and unlabeled sets are constructed as follows:
\begin{equation}\label{eq:train_dataset}
   \begin{split}
       D_{tx}^{(A)} &= \left\{(\mathbf{x}_i,y_i, w_i^{(V)})\middle|\, w_i^{(V)} \geq \tau_w, 
           (\mathbf{x}_i,y_i)\in D_t \;  \text{and} \; w_i^{(V)} \in \mathcal{W}^{(V)}\right\}, \\
       D_{tu}^{(A)} &= \left\{\mathbf{x}_i\middle|\, w_i^{(V)} < \tau_w, 
           (\mathbf{x}_i,y_i)\in D_t \;  \text{and} \; w_i^{(V)} \in \mathcal{W}^{(V)}\right\}, \\
       D_{tx}^{(V)} &= \left\{(\mathbf{x}_i,y_i, w_i^{(A)})\middle|\, w_i^{(A)} \geq \tau_w, 
           (\mathbf{x}_i,y_i)\in D_t \;  \text{and} \; w_i^{(A)} \in \mathcal{W}^{(A)}\right\},\\
   \end{split}
\end{equation}
Samples with a clean probability greater than or equal to the threshold $\tau_w$ are considered labeled samples with reliable labels. For samples with a probability below $\tau_w$,  their original labels are discarded, and they are treated as unlabeled data.
A key distinction from existing symmetric co-teaching methods~\cite{Han_2018_Co-teaching, Li_2020_DivideMix} is our asymmetric training strategy. While net $A$ is trained semi-supervisedly on both labeled ($D_{tx}^{(V)}$) and unlabeled  ($D_{tu}^{(A)}$)samples, net $V$ is  trained on only the labeled samples ($D_{tx}^{(A)}$). This design is based on the rationale that for  the pretrained ViT, which maintains high prediction accuracy from the beginning of training, the risk of performance degradation from using unreliable labels outweighs the potential benefit of training on more samples.

\subsubsection{\textbf{Phase 2}: Pseudo-labeling}
Following the DivideMix procedure, this phase generates refined pseudo-labels, $\bar{\mathbf{y}}_i$, for the training samples. For labeled samples, the pseudo-label is created by correcting the original label, $y_i$, using the training network’s prediction $p_{\theta_l}(\mathbf{x}_i)$, and the estimated clean probability, $w_i$. For unlabeled samples, new pseudo-labels are generated by combining the predictions from both the training network and its peer model.

\subsubsection{\textbf{Phase 3}: Mixup augmentation}
To enhance the models’ robustness to noisy samples, we employed Mixup augmentation~\cite{Zhang_2017_MixUp}, which generates new training data by creating linear interpolations of randomly selected image pairs ($(\mathbf{x}_i, \mathbf{x}_j)$) and their corresponding pseudo-label pairs ($(\bar{\mathbf{y}}_i, \bar{\mathbf{y}}_j)$). This process results in new augmented images, $\hat{\mathbf{x}}_i$, and their associated mixed labels, $\hat{\mathbf{y}}_i$. For net $A$, which is trained semi-supervisedly, the sample pairs used for Mixup are drawn from the combined pool of both its labeled and unlabeled datasets.

\subsubsection{\textbf{Phase 4}: Training}
In the final phase, each network is trained using the samples generated by Mixup augmentation. The training objective is composed of several loss terms. For the labeled samples $(\hat{\mathbf{x}}_{tx}, \hat{\mathbf{y}}_{tx}) from either ({D}_{tx}^{(A)} \text{ or } {D}_{tx}^{(V)})$, we use the standard cross-entropy loss:
\begin{equation}\label{eq:train labels}
   \mathcal{L}_x = -\hat{\mathbf{y}}_{tx}^{\top}\log(p_{\theta_l}(\hat{\mathbf{x}}_{tx})). 
\end{equation}
For the augmented unlabeled samples, $\{(\hat{\mathbf{x}}_{tu}, \hat{\mathbf{y}}_{tu}) \in {D}_{tu}^{(A)}\}$, the consistency loss is defined as:
\begin{equation}\label{eq:train unlabels}
   \mathcal{L}_u = ||\hat{\mathbf{y}}_{tu} - p_{\theta_l}(\hat{\mathbf{x}}_{tu})||^2_2.
\end{equation}
The final training objectives for net $A$ ($\mathcal{L}^{(A)}$) and net $V$ ($\mathcal{L}^{(V)}$) combine these loss terms and reflect the framework's asymmetric design:
\begin{equation}\label{eq:loss}
   \begin{split}
      \mathcal{L}^{(A)} &= \mathcal{L}_x + \lambda_u\mathcal{L}_u + \mathcal{L}_{reg}, \\
       \mathcal{L}^{(V)} &= \mathcal{L}_x + \mathcal{L}_{reg},
   \end{split}
\end{equation}
where $\mathcal{L}_{reg}$ is the regularization term adopted from DivideMix, and $\lambda_u$ is a hyperparameter that weights the contribution of the unlabeled consistency loss.

\begin{algorithm}[t]
\SetAlgoLined
\caption{Main training algorithm for the ACD-U framework}
\label{alg:poposed}
\KwIn{dataset $D$, $E_{warmup}$, $E_{start}$, $E_{encoder}$, $E_{UP}$, $E_{UD}$, $B$, $B_u$, $\tau_w$, $\lambda_u$, $T_{unl}$, $p_{low}$, $p_{drop}$}
\KwOut{trained networks $\theta_A$, $\theta_V$}

\While{$k \leq \text{MaxEpoch}$}{
   \If{$k \leq E_{warmup}$}{
        \tcp{Warmup period: stabilize initial learning}
        Perform self-supervised learning for net $V$ without labels\;
        Perform supervised learning for net $A$ with dataset $D$\;
   }
   \Else{
        \If{$k \geq E_{start}$ \textbf{and} $k \,\%\, E_{UP} = 0$}{
            \tcp{Select unlearning samples every $E_{UP}$ epochs after $E_{start}$}
           $D_u^{(A)}, D_u^{(V)}, D_t = \text{UnlearningSetup}(D, \theta_A, \theta_V, k, p_{low}, p_{drop})$\;
       }
       \If{$k \geq E_{start}$ \textbf{and} $k \,\%\, E_{UP} \leq E_{UD}$}{
            \tcp{Apply unlearning for $E_{UD}$ epochs after each selection}
           $\text{ApplyUnlearning}(D_u^{(A)}, D_u^{(V)}, \theta_A, \theta_V, B_u, T_{unl})$\;
       }
       \tcp{Perform asymmetric co-teaching}
       $\text{AsyCoT}(D_t, \theta_A, \theta_V, k, B, \tau_w, \lambda_u, E_{encoder})$\;
   }
}
\end{algorithm}

\begin{algorithm}[h]
\SetAlgoLined
\caption{Unlearning setup procedure}
\label{alg:unlearning_setup}
\KwIn{dataset $D$, network parameters $\theta_A$, $\theta_V$, epoch $k$, $p_{low}$, $p_{drop}$}
\KwOut{$D_u^{(A)}$, $D_u^{(V)}$, $D_t$}

\For{$l \in \{A, V\}$}{
   $D_u^{(l)} = \text{UnlearningSS}(l, \theta_l, D, \mathcal{L}_s^{(k-E_{UP})}(\cdot; \theta_l), p_{low}, p_{drop})$\;
   $\theta_{ref}^{(l)} = \theta_l$\;
}
$D_t = D\setminus (D_u^{(A)} \cup D_u^{(V)})$\;
\end{algorithm}

\begin{algorithm}[h]
\SetAlgoLined
\caption{Procedure for applying the unlearning loss}
\label{alg:apply_unlearning}
\KwIn{$D_u^{(A)}$, $D_u^{(V)}$, network parameters $\theta_A$, $\theta_V$, $B_u$, $T_{unl}$}
\KwOut{updated $\theta_A$, $\theta_V$}

\For{$l \in \{A, V\}$}{
   \For{mini-batch $\mathcal{B}_u$ from $D_u^{(l)}$ with size $B_u$}{
       $\mathcal{L}_{unl} = -T_{unl}^2\sum_{\mathbf{x}_u \in \mathcal{B}_u} D_{KL}(p_{\theta_{ref}^{(l)}}(\mathbf{x}_u)\| p_{\theta_l}(\mathbf{x}_u))$\;
       Update $\theta_l$ using $\mathcal{L}_{unl}$\;
   }
}
\end{algorithm}

\begin{algorithm}[h]
\SetAlgoLined
\caption{Asymmetric co-reaching (AsyCoT) procedure}
\label{alg:asymmetric_coteaching}
\KwIn{dataset $D_t$, network parameters $\theta_A$, $\theta_V$, epoch $k$, $B$, $\tau_w$, $\lambda_u$, $E_{encoder}$}
\KwOut{updated $\theta_A$, $\theta_V$}

Compute $\mathcal{W}^{(A)}$, $\mathcal{W}^{(V)}$ using GMM on $D_t$\;
Construct $D_{tx}^{(A)}$, $D_{tu}^{(A)}$, $D_{tx}^{(V)}$ using $\mathcal{W}^{(A)}$, $\mathcal{W}^{(V)}$ and threshold $\tau_w$\;

\tcp{Train net $A$ with labeled and unlabeled samples}
\For{mini-batch from $D_{tx}^{(A)}$, $D_{tu}^{(A)}$ with size $B$}{
   Apply Mixup and compute $\mathcal{L}^{(A)} = \mathcal{L}_x + \lambda_u\mathcal{L}_u + \mathcal{L}_{reg}$\;
   Update $\theta_A$ using $\mathcal{L}^{(A)}$\;
}

\tcp{Train net $V$ with labeled samples only}
\For{mini-batch from $D_{tx}^{(V)}$ with size $B$}{
   Apply Mixup and compute $\mathcal{L}^{(V)} = \mathcal{L}_x + \mathcal{L}_{reg}$\;
   \If{$k < E_{encoder}$}{
       Update only classification head of $\theta_V$\;
   }
   \Else{
       Update both encoder and classification head of $\theta_V$\;
   }
}
\end{algorithm}

\subsection{Training algorithm}\label{subsec:training_algorithm}
The overall learning algorithm for ACD-U, detailed in Algorithm~\ref{alg:poposed}, is structured into three distinct stages:

\noindent \textbf{Warmup Period (Epochs 1 to $E_{warmup}$)}:
To ensure a stable start to training, this initial phase is dedicated to network stabilization. During this warmup period, net $A$ is trained using normal supervised learning, whereas net $V$ undergoes self-supervised learning without using any labels.

\noindent \textbf{Preparation Period (Epochs $E_{warmup}+1$ to $E_{start}$)}:
In this stage, the ACD framework is applied using the entire dataset $D$. The unlearning mechanism  is not yet active. To further stabilize the process, the encoder of net $V$ remains frozen until epoch $E_{encoder}$; only its classification head is trained before this point.

\noindent \textbf{Unlearning Execution Period (Epoch $E_{start}$ onwards)}:
The complete ACD-U framework is activated. The process of selecting forgetting targets and applying the unlearning loss is executed periodically at intervals of $E_{UP}$ epochs. This occurs in parallel with the continued ACD training, which now proceeds using the filtered training dataset $D_t$.

This multi-stage design promotes stable learning by systematically building model capacity. The warmup period ensures robust initialization, the preparation period allows the models to learn general features from all available data, and the final stage introduces the periodic unlearning mechanism to gradually remove the influence of noisy samples. By carefully determining when to begin fine-tuning the ViT encoder, the process further avoids early instability and refines the model’s representations.

\subsection{Position of the proposed method}
ACD-U's unique integration of CLIP, unlearning, co-teaching, and a pretrained VLM overcomes the critical issue of irreversible error memorization in noisy-label learning. Standard co-teaching methods are vulnerable to confirmation bias, particularly when both networks mistakenly agree that a noisy sample is clean---a documented problem in early training~\cite{takeda_2021_IEEE}. Once such an agreement occurs, the error becomes embedded and persists, as existing refinements~\cite{xiao_2022_promix, Zhang_2023_Rankmatch} lack a mechanism to correct past selection errors.

Furthermore, our approach is designed to exploit the heterogeneous learning dynamics of different model architectures. Pretrained ViTs and randomly initialized CNNs memorize noisy samples at different rates, a distinction that most existing methods ignore. Pretrained models maintain stable predictions but risk overfitting to noisy labels during fine-tuning, whereas CNNs improve gradually but accumulate errors progressively. Current approaches either treat pretrained models as static validators~\cite{Feng_2024_Clipcleaner} or train different architectures with identical symmetric strategies~\cite{ahn_2023_arXiv, wang_2025_arXiv}. By contrast, we leverage these complementary behaviors through asymmetric training: the ViT trains only on high-confidence clean samples, providing a stable learning signal, whereas the CNN uses SSL to learn from both clean and noisy data.

ACD-U also makes machine unlearning a viable tool for the LNL domain. Applying unlearning to noisy labels requires solving a key challenge: identifying which samples have been incorrectly memorized during training. This is distinct from privacy applications, where the samples to be forgotten are known beforehand~\cite{Kurmanji_2023_SCRUB, bourtoule_2021_sisa}. While SAP~\cite{sangamesh2025sap} also applies unlearning, it is limited to pretrained models and does not address training from scratch. We solve this discovery problem with a novel selection mechanism that combines loss trajectory analysis with external CLIP consistency checks (Eqs.~\eqref{eq:low_loss_quantile}--\eqref{eq:unlearning_selection}), allowing us to identify memorized errors without prior knowledge of the noisy samples.

Ultimately, these components form an interdependent error correction system. The pretrained ViT guides the CNN’s learning, helps identify samples the ViT should avoid, and the unlearning mechanism corrects errors that neither architecture can resolve alone. This integrated feedback loop distinguishes our framework from methods that apply sequential techniques~\cite{Feng_2024_NoiseBox} or use fixed, pretrained components~\cite{Feng_2024_Clipcleaner}. The temporal gap created by the ViT’s initial stability and the CNN’s gradual adaptation enables this novel error identification and correction. Instead of focusing solely on preventing errors through better initial selection, ACD-U actively detects and removes them throughout the training process. This addresses the core limitation of existing methods, which cannot recover once a sample has been misclassified. These key distinctions are summarized in Table~\ref{tb:technical_gaps}.

\begin{table*}[t]
\centering
\caption{Summary of key technical gaps in existing LNL methods and the corresponding solutions provided by the ACD-U framework.}
\label{tb:technical_gaps}
\resizebox{\textwidth}{!}{%
\begin{tabular}{p{3.5cm}p{5cm}p{4.5cm}p{4cm}}
\toprule
\textbf{Technical Gap} & \textbf{Problem in Existing Methods} & \textbf{Representative Methods} & \textbf{ACD-U Solution} \\
\midrule
\textbf{Irreversible Error Embedding} & When both networks misclassify a noisy sample as clean, error becomes permanent with no correction mechanism & Co-teaching~\cite{Han_2018_Co-teaching}, DivideMix~\cite{Li_2020_DivideMix}, ProMix~\cite{xiao_2022_promix}, RankMatch~\cite{Zhang_2023_Rankmatch} & Selective unlearning enables post-hoc removal of memorized errors through loss trajectory analysis \\
\midrule
\textbf{Uniform Training of Heterogeneous Models} & Different architectures trained identically, ignoring their distinct memorization patterns & CLIPCleaner~\cite{Feng_2024_Clipcleaner} (static validator), TURN~\cite{ahn_2023_arXiv} (frozen ViT extractor) & Asymmetric training: ViT uses clean samples only, CNN uses SSL on all data \\
\midrule
\textbf{Unknown Forgetting Targets} & Standard unlearning assumes prior knowledge of problematic samples, incompatible with noisy-label discovery during training & SCRUB~\cite{Kurmanji_2023_SCRUB}, SAP~\cite{sangamesh2025sap} (pretrained only) & Dynamic identification via loss changes + CLIP consistency checks (Eqs.~\ref{eq:low_loss_quantile}-\ref{eq:unlearning_selection}) \\
\midrule
\textbf{Independent Component Operation} & Techniques applied sequentially without feedback between components & NoiseBox~\cite{Feng_2024_NoiseBox} (sequential), CLIPCleaner~\cite{Feng_2024_Clipcleaner} (fixed CLIP) & Interdependent system: ViT guides CNN, CNN informs ViT, unlearning corrects both \\
\bottomrule
\end{tabular}}
\end{table*}

\section{Experiments}\label{sec:04}
This section details the experiments conducted to validate our proposed method. Our evaluation is guided by the following research questions (RQs):
\begin{itemize}
   \item \textbf{RQ1}: Does ACD-U demonstrate superior performance than existing methods in various noise environments?
   \item \textbf{RQ2}: How do selective unlearning and ACD each contribute to performance improvement?
   \item \textbf{RQ3}: What are the optimal design choices and parameter settings to maximize the performance of the proposed method?
\end{itemize}

\begin{table*}[ht]
\centering
\caption{Summary of the comparison methods evaluated on each dataset.}
\label{tb:comparison}
\begin{tabular}{lccccc}
\toprule
\textbf{Method} & \textbf{CIFAR-10/-100} & \textbf{CIFAR-N} & \textbf{Red Mini-ImageNet} & \textbf{Clothing1M} & \textbf{WebVision} \\
\midrule
DivideMix~\cite{Li_2020_DivideMix} & \checkmark & \checkmark & \checkmark & \checkmark & \checkmark \\
LongReMix~\cite{cordeiro_2023_LongReMix} & \checkmark &  & \checkmark & \checkmark & \checkmark \\
ProMix~\cite{xiao_2022_promix} & \checkmark & \checkmark & & \checkmark & \\
RankMatch~\cite{Zhang_2023_Rankmatch} & \checkmark & & & \checkmark & \checkmark \\
SV-Learner~\cite{Liang_2024_SV-Learner} & \checkmark & & & \checkmark & \checkmark \\
NoiseBox+SS-KNN ~\cite{Feng_2024_NoiseBox} & \checkmark & & \checkmark & \checkmark & \checkmark \\
Semi-RML++ ~\cite{li_2025_RML} & \checkmark & \checkmark & & \checkmark & \checkmark \\
CLIPCleaner~\cite{Feng_2024_Clipcleaner} & \checkmark & & \checkmark & \checkmark & \checkmark \\
LSL~\cite{kim_2024_LSL} & & \checkmark & \checkmark & & \checkmark \\
\bottomrule
\end{tabular}
\begin{tablenotes}
    \small
   \item Note: A \checkmark indicates that performance values were taken directly from the method's original paper, following the standard protocol for that dataset. For DivideMix, the results on CIFAR-N and Red Mini-ImageNet, which were not reported in the original publications, are cited from other works~\cite{xiao_2022_promix, cordeiro_2023_LongReMix}.
  \end{tablenotes}
\end{table*}

\begin{table*}[t]
 \centering
 \caption{Hyperparameter settings for the proposed method across all experimental datasets.}
 \label{tb:hyperparameters}
 \resizebox{\textwidth}{!}{%
  \begin{tabular}{lccccccc}
   \toprule
   \multicolumn{1}{c}{\textbf{Parameter}} & \multicolumn{7}{c}{\textbf{Dataset}} \\
   \cmidrule(lr){2-8}
   & CIFAR-10 & CIFAR-100 & CIFAR-10N & CIFAR-100N & Red Mini-ImageNet & WebVision & Clothing1M \\
   \midrule 
   Architecture (net $A$) & PreAct ResNet18 & PreAct ResNet18 & PreAct ResNet18 & PreAct ResNet18 & PreAct ResNet18 & ResNet-50 & ResNet-50 \\
   Architecture (net $V$) & ViT & ViT & ViT & ViT & ViT & ViT & ViT \\
   Optimizer & SGD & SGD & SGD & SGD & SGD & SGD & SGD \\
   Momentum & 0.9 & 0.9 & 0.9 & 0.9 & 0.9 & 0.9 & 0.9 \\
   Weight Decay & 0.0005 & 0.0005 & 0.0005 & 0.0005 & 0.0005 & 0.001 & 0.001 \\
   Batch Size $B$ & 128 & 128 & 128 & 128 & 128 & 64 & 64 \\
   Initial Learning Rate (net $A$) & 0.02 & 0.02 & 0.02 & 0.02 & 0.02 & 0.01 & 0.002 \\
   Initial Learning Rate (net $V$) & 0.002 & 0.002 & 0.002 & 0.002 & 0.002 & 0.002 & 0.002 \\
   Epoch to decay learning rate by 1/10 & 150 & 150 & 150 & 150 & 150 & 50 & 40 \\
   Toal Epoch & 300 & 300 & 300 & 300 & 300 & 100 & 80 \\
   $E_{warmup}$ & 10 & 30 & 10 & 30 & 30 & 1 & 1 \\
   $E_{start}$ & 60 & 60 & 60 & 60 & 60 & 40 & 40 \\
   $E_{encoder}$ & 50 & 50 & 50 & 50 & -- & 20 & 20 \\
   $\lambda_u$ & $\star$ & $\star$ & 1 & 1 & 25 & 0 & 0 \\
   $p_{low}$ & 0.05 & 0.05 & 0.05 & 0.05 & 0.05 & 0.05 & 0.05 \\
   $p_{drop}$ &  0.2 & 0.2 & 0.2 & 0.2 & 0.2 & 0.2 & 0.2 \\
   Batch Size for Unlearning $B_u$ & 512 & 512 & 512 & 512 & 512 & 256 & 128 \\
   $T_{unl}$ & 0.05 & 0.05 & 0.05 & 0.05 & 0.05 & 0.05 & 1.0 \\
   \bottomrule
  \end{tabular}}
  \begin{tablenotes}
    \small
    \item Note: $\star$ indicates that $\lambda_u$ varies by noise rate and uses the same values as those reported in~\cite{Li_2020_DivideMix}.
  \end{tablenotes}
\end{table*}

\subsection{Experimental conditions}
\subsubsection{Datasets}
\noindent \textbf{Noise models.} We evaluated our method under three common types of label noise: (i) class-independent \textbf{Symmetric noise}~\cite{van_2015_NeurIPS}, (ii) class-dependent \textbf{Asymmetric noise}~\cite{Patrini_2017_IEEE}, and (iii) instance-dependent \textbf{Instance-dependent noise}~\cite{dawson_2023_beyond}. Let $C$ be the number of classes, $\eta$ be the noise rate, $y\in\{1,\dots,C\}$ be the clean label, $\tilde{y}\in\{1,\dots,C\}$ be the observed label, and $i,j\in\{1,\dots,C\}$ be class subscripts. The class-conditional label transition probability is defined as $T_{ij}=P(\tilde{y}=j\mid y=i)$, where each row forms a probability distribution ($\sum_{j}T_{ij}=1,\ T_{ij}\ge 0$). For symmetric noise, the transition probabilities are
\begin{equation}
   T_{ii}=1-\eta,\qquad T_{ij}=\frac{\eta}{C-1}\ (j\neq i).
\end{equation}
For asymmetric noise, where class $i$ is flipped to a single confusable class $j^\star(i)$, the probabilities are
\begin{equation}
   T_{ii}=1-\eta,\qquad T_{i\,j^\star(i)}=\eta,\qquad T_{ij}=0\ (j\notin\{i,j^\star(i)\}).
\end{equation}
The generation mechanism for instance-dependent noise is expressed as $T_{ij}(x)=P(\tilde{y}=j\mid y=i,x)$, where the corruption depends on the instance features $x$. As this mechanism is typically unknown, we evaluate it using real-world datasets such as Clothing1M, which inherently contain this type of noise.

\noindent \textbf{Benchmarks and protocols.}
To systematically evaluate performance, we used a comprehensive suite of datasets. For controlled experiments, we injected synthetic noise in the CIFAR series~\cite{Krizhevsky_2009}. To test robustness against real-world human annotation errors, we used CIFAR-N~\cite{wei_2022_CIFAR_N}. We also assessed performance on large-scale, near-real-world noise distributions using WebVision~\cite{li_2017_webvision}, Clothing1M~\cite{xiao_2015_clothing}, and Red Mini-ImageNet~\cite{jiang_2020_PMLR}. Our experimental setup, including input resolutions, preprocessing, and data splits, adhered to the standard protocols for each dataset. This ensured fair comparison and reproducibility, as these benchmarks are widely used in recent leading methods~\cite{Li_2020_DivideMix,cordeiro_2023_LongReMix,Zhang_2023_Rankmatch,xiao_2022_promix,Liang_2024_SV-Learner,Feng_2024_NoiseBox,Feng_2024_Clipcleaner,li_2025_RML}.
\begin{enumerate}
 \item \textbf{CIFAR--10 / --100}: 
  CIFAR-10 (10 classes) and CIFAR-100 (100 classes) both  contain 50K training and 10K test images ($32\times32$). As these datasets contain accurate labels, we artificially introduced symmetric noise at  20\%, 50\%, 80\%, 90\% rates and asymmetric noise, which flips labels to visually similar classes (e.g., truck $\rightarrow$ automobile). For asymmetric noise, we used a 40\% rate, as class identification becomes difficult above 50\% ~\cite{Huang_2023_CVPR_TCL}. We report the highest accuracy across all epochs (\textbf{Best}) and the average accuracy over the final 10 epochs (\textbf{Last}).

 \item \textbf{CIFAR--N}:
 This benchmark comprises CIFAR-10N and CIFAR-100N, which contain real-world noisy labels from human annotators on Amazon Mechanical Turk. Following~\cite{xiao_2022_promix}, we evaluated three CIFAR-10N noise types with rates of 9.03\% (Aggregate), 17.23\% (Rand1), and 40.21\% (Worst), and on the CIFAR-100N “Noisy-Fine” set with a 40.20\% noise rate.

 \item \textbf{WebVision}: 
 The WebVision dataset contains 2.4M images with noisy labels scraped from the web across the 1000 classes of ImageNet ILSVRC12~\cite{Mehrasa_2009_ImageNet}. Each image is resized to $256\times256$. Following the protocol in~\cite{Li_2020_DivideMix}, we used the “mini-WebVision” subset, which includes the first 50 classes. Models were evaluated for \textbf{top-1} and \textbf{top-5} accuracy on both the WebVision and ImageNet ILSVRC12 validation sets.

 \item \textbf{Clothing1M}:
 This large-scale dataset, collected from online shopping sites, contains approximately 1M training images across 14 clothing categories, with an estimated instance-dependent noise rate of 38.5\%~\cite{Song_2020_prestopping}. It is a key benchmark for instance-dependent noise~\cite{dawson_2023_beyond}. Following the protocol from~\cite{Li_2020_DivideMix}, we evaluated the model’s performance on the validation dataset at the end of each epoch. The model weights from the epoch that achieved the highest validation accuracy  are then saved. Finally, these best-performing weights are used to conduct the final evaluation on the test set.

 \item \textbf{Red Mini--ImageNet}:
 This dataset is part of the Controlled Noisy Web Labels (CNWL) dataset~\cite{jiang_2020_PMLR}, comprising 100 classes with 50K training and 5K test images. The CNWL dataset contains web-sourced images where label correctness was determined by a majority vote of 3--5 annotators. The benchmark simulates varying noise levels by replacing $p$\% of the clean data with known mislabeled samples. Following~\cite{Xu_2021_CVPR}, we used noise rates of $p=20,40,60,80$\%, the $32\times32$ resized training images published by~\cite{Xu_2021_CVPR}, and test images from ImageNet ILSVRC12~\cite{Mehrasa_2009_ImageNet}.
\end{enumerate}

\subsubsection{Comparison methods}
We systematically compared ACD-U against representative and state-of-the-art frameworks in LNL. The selected methods are organized into four groups based on their core approach: (A) Co-teaching-based sample selection + SSL, (B) selection and reweighting based on self-verification and confidence estimation, (C) label correction using a aVLM (CLIP) as an external teacher, and (D) enhanced learning frameworks that incorporate noise assumptions. The comparisons were performed against publicly reported results from these methods, following the standard evaluation protocol for each dataset to ensure fairness and consistency.

\begin{itemize}
 \item[(A)] \textbf{DivideMix} ('20)~\cite{Li_2020_DivideMix}: A foundational method that combines co-teaching with MixMatch, using a GMM to estimate clean and noisy sample probabilities.
 \textbf{LongReMix} ('22)~\cite{cordeiro_2023_LongReMix}: An extension that optimizes learning schedules and mixing strategies to stabilize long-term training.
 \textbf{ProMix} ('22)~\cite{xiao_2022_promix}: Balances selection accuracy and generalization using a curriculum-based approach and Mixup.
 \textbf{RankMatch} ('23)~\cite{Zhang_2023_Rankmatch}: Improves the quality of pseudo-labels by enforcing rank consistency.

 \item[(B)] \textbf{SV-Learner} ('24)~\cite{Liang_2024_SV-Learner}: Self-verifies the reliability of samples based on prediction consistency and adjusts their contribution to the learning objective. 
 \textbf{LSL} ('24)~\cite{kim_2024_LSL}: A method specialized for human-generated noise (e.g., in CIFAR-N) that adapts to annotation variance through label self-learning.

 \item[(C)] \textbf{CLIPCleaner} ('24)~\cite{Feng_2024_Clipcleaner}: Uses a pretrained CLIP model as an external teacher to correct noisy training labels. This serves as a strong baseline owing to its similar use of CLIP.

 \item[(D)] \textbf{NoiseBox+SS-KNN} ('24)~\cite{Feng_2024_NoiseBox}:  A general framework that enhances existing selection methods by integrating subset expansion and using SS-KNN to leverage feature space relationships.
 \textbf{Semi-RML++} ('25)~\cite{li_2025_RML}: A semi-supervised extension of robust meta-learning with enhanced regularization and memory-consistency constraints.

\end{itemize}

Table~\ref{tb:comparison} summarizes the comparison methods used for each dataset. The selection of methods varies to respect the standard protocols of each benchmark and ensure fairness by using the originally reported performance values.

\subsubsection{Implementation Details}
For all experiments, the pretrained ViT image encoder from CLIP~\cite{Radford_2021_CLIP} was used as the backbone for net $V$. For net $A$, we adopted the standard CNN architectures specified by the established benchmark protocols for each dataset (see Table~\ref{tb:hyperparameters}). Our focus was to demonstrate the effectiveness of the ACD framework by combining a pretrained ViT with a conventional CNN, rather than performing an exhaustive search over all possible architectural combinations. While other pretrained architectures could potentially serve as net $V$---such as Swin Transformer~\cite{liu2021swin}, DeiT~\cite{touvron2021deit}, ConvNeXt~\cite{liu2022convnext}, or other vision foundation models such as DINOv2~\cite{oquab2023dinov2} and SAM~\cite{kirillov2023sam}--- a systematic exploration of these alternatives would require extensive computational resources and is beyond the scope of this study. We selected CLIP's ViT specifically for two key reasons. First, its strong zero-shot capabilities are essential for our unlearning sample-selection mechanism (Eq.~\eqref{eq:CLIP_selection}), and second, its vision-language alignment offers unique advantages for noise detection that purely visual models cannot provide.

For the unlearning sample-selection phase, we used an independent, fixed CLIP-V/32 model~\cite{clip_pre_train}. The zero-shot predictions were generated using the class name template ``a photo of a \{class\_name\}''. Additionally, because the ViT backbone lacks a final classification layer, we followed~\cite{Huang_2023_CVPR_TCL} and added a two-layer multi-layer perceptron as a classification head. To manage computational costs, all CLIP-based models were operated using half-precision (FP16) floating-point numbers. The input image size for net $V$ was unified to $224\times224$ to match the ViT's pretrained resolution. To stabilize the initial learning phases, we froze the encoder of net $V$ for the first $E_{encoder}$; this step was omitted for Red Mini-ImageNet. During the warmup period, we adopted the self-supervised approach of training CLIP without labels~\cite{laroudie_2023_arXiv}. However, for the large-scale WebVision and Clothing1M datasets, we provided a weak supervisory signal during warmup by using pseudo-labels created by averaging CLIP's zero-shot predictions with the original noisy labels in a 1:1 ratio.

Table~\ref{tb:hyperparameters} lists the hyperparameters used in the experiments. The architecture for net $A$ in each experiment was selected to match the standard protocol for that dataset. To ensure fair comparisons, we adopted the same core hyperparameters as prior works~\cite{Li_2020_DivideMix,xiao_2022_promix}. Hyperparameters unique to the proposed method were determined empirically, and their sensitivity was analyzed through parameter analysis.

\begin{table*}[t]
 \centering
 \caption{Test accuracies (\%) on the CIFAR-10 and CIFAR-100 datasets with synthetic symmetric and asymmetric noise. “Best” indicates the highest accuracy achieved across all training epochs, and “Last” is the average accuracy over the final 10 epochs, based on~\cite{Li_2020_DivideMix, xiao_2022_promix, Zhang_2023_Rankmatch}.}
 \label{tb:results_CIFAR10_100}
 \newcolumntype{C}{>{\centering\arraybackslash}X}
 \begin{tabularx}{\textwidth}{l l *{9}{C}}
  \toprule
  \multicolumn{2}{l}{Dataset} & \multicolumn{5}{c}{\textbf{CIFAR-10}} & \multicolumn{4}{c}{\textbf{CIFAR-100}}\\
  \cmidrule(lr){3-7} \cmidrule(lr){8-11}
  \multicolumn{2}{l}{Noise type} & \multicolumn{4}{c}{Sym.} & Asym. & \multicolumn{4}{c}{Sym.}\\
  \multicolumn{2}{l}{Method/Noise ratio} & 20\% & 50\% & 80\% & 90\% & 40\% & 20\% & 50\% & 80\% & 90\%\\
  \midrule
  \multirow{2}{*}{DivideMix~\cite{Li_2020_DivideMix}} 
           & Best & 96.1 & 94.6 & 93.2 & 76.0 & 93.4 & 77.3 & 74.6 & 60.2 & 31.5\\
           & Last & 95.7 & 94.4 & 92.9 & 75.4 & 92.1 & 76.9 & 74.2 & 59.6 & 31.0\\
    \multirow{2}{*}{LongReMix~\cite{cordeiro_2023_LongReMix}}	
            & Best & 96.3 & 95.1 & 93.8 & 79.9 & 94.7 & 77.9 & 75.5 & 62.3 & 34.7 \\
            & Last & 96.0 & 94.8 & 93.3 & 79.1 & 94.3 & 77.5 & 74.9 & 61.7 & 30.7  \\
    \multirow{2}{*}{ProMix~\cite{xiao_2022_promix}} 
           & Best & \textbf{97.7} & \textbf{97.4} & \underline{95.5} & 93.4 & \textbf{96.6} & \underline{82.6} & \underline{80.1} & \underline{69.4} & 42.9\\
           & Last & \textbf{97.6} & \textbf{97.3} & 95.1 & 91.1 & \textbf{96.5} & \underline{82.4} & 79.7 & 69.0 & 42.7\\ 
    \multirow{2}{*}{RankMatch~\cite{Zhang_2023_Rankmatch}} 
           & Best & 96.5 & 95.6 & 94.5 & 92.6 & 94.7 & 79.5 & 77.9 & 67.6 & 50.6\\
           & Last & 96.4 & 95.4 & 94.2 & 92.1 & 94.4 & 79.3 & 77.6 & 67.2 & 49.9\\
    \multirow{1}{*}{SV-Learner~\cite{Liang_2024_SV-Learner}} 
            & Best & 96.9 & 96.3 & 94.8 & 92.4 & 95.3 & 81.2 & 78.6 & 65.2 & 49.6\\
    \multirow{1}{*}{NoiseBox+SS-KNN~\cite{Feng_2024_NoiseBox}} 
            & Last  & 96.6 & 96.2 & \underline{95.7} & \textbf{95.4} & \underline{96.0} & 79.4 & 77.4 & \underline{72.8} & \textbf{67.1}\\
    \multirow{1}{*}{Semi-RML++ ~\cite{li_2025_RML}} 
            & Last & \underline{97.1} & \underline{96.8} & \textbf{95.8} & - & - & 81.5 & \underline{80.6} & 70.4 & - \\
    \multirow{1}{*}{CLIPCleaner~\cite{Feng_2024_Clipcleaner}} 
            & Best & 95.9 & 95.7 & 95.0 & \textbf{94.2} & 94.9 & 78.2 & 75.2 & 69.7 & \underline{63.1}\\
\midrule
  \multirow{2}{*}{ACD-U}
    & Best &\underline{97.2} & \underline{96.8} & \textbf{95.6} & \underline{93.9} & \underline{95.7} & \textbf{83.3} & \textbf{81.6} & \textbf{74.4} & \textbf{66.5}\\
  & Last & 97.0 & 96.6 & 95.4 & \underline{93.3} & 94.4 & \textbf{83.0} & \textbf{81.2}& \textbf{73.3} & \underline{63.4}\\
  \bottomrule
 \end{tabularx}
\end{table*}

\subsection{Main Results (RQ1)}\label{sec:04:02}
To answer RQ1, we evaluated ACD-U on a diverse range of datasets featuring both synthetic (Section~\ref{sec:04:02:01}) and real-world (Section~\ref{sec:04:02:02}) noise. The experiments analyzed performance across three key axes: noise properties (symmetric, asymmetric, and instance-dependent), noise rates (20\%---90\%), and overall dataset scale and complexity.

\subsubsection{Performance on synthetic-noise environments}\label{sec:04:02:01}
The results, listed in Table~\ref{tb:results_CIFAR10_100}, indicate that our method demonstrates strong performance on the CIFAR-10 and CIFAR-100  datasets with synthetic noise. On the more challenging CIFAR-100 benchmark, ACD-U outperforms existing methods under most conditions. In the extreme Sym.90\% setting, it achieves a 35\% relative improvement over DivideMix, demonstrating that the selective unlearning and ACD components are highly effective for challenging, multi-class classification tasks. Even at a moderate noise rate (Sym.50\%), ACD-U outperforms strong baselines, such as ProMix and Semi-RML++, by more than 0.6\%.

On CIFAR-10, the performance gains are more modest. At low noise rates, our method’s accuracy is within 0.5--0.6\% of ProMix, whereas at high noise rates, NoiseBox+SS-KNN shows superior performance. We attribute this to the relative simplicity of the 10-class classification task, where existing methods already achieve accuracies above 90\%, leaving little room for significant improvement. This suggests that existing methods are already highly optimized for simple tasks.

When compared against other methods that also leverage CLIP, ACD-U consistently outperforms CLIPCleaner. This advantage is particularly pronounced on CIFAR-100 in low-to-medium noise regimes, where it achieved more than 5\% higher accuracies. This significant difference is attributed to our core design: while CLIPCleaner uses CLIP as a static, external validator, ACD-U integrates the pretrained ViT as a learnable component that fine-tunes to the specific task. In the asymmetric noise setting, our method outperforms DivideMix by 2.3\% but is surpassed by ProMix by 0.9\%.

\begin{table}[t]
  \centering
  \caption{Test accuracies (\%) on the real-world CIFAR-10N and CIFAR-100N benchmarks.}
  \label{tb:CIFAR-N}
  \begin{tabular}{llcccc}
  \toprule
  \multicolumn{1}{l}{Dataset}
  
  & \multicolumn{3}{c}{\textbf{CIFAR-10N}}
  & \multicolumn{1}{c}{\textbf{CIFAR-100N}} \\
  \cmidrule(lr){2-4} \cmidrule(lr){5-5}
  Method      & aggre & random & worst & noisy \\
  \midrule
  DivideMix~\cite{Li_2020_DivideMix}    & 95.01 & 95.16 & 92.56 & 71.13 \\
  ProMix~\cite{xiao_2022_promix}      & \textbf{97.65} & \textbf{97.39} & \textbf{96.34} & 73.79 \\
  LSL~\cite{kim_2024_LSL}          & - & - & 94.57 & 74.46 \\
  Semi-RML++~\cite{li_2025_RML}          & 96.84 & 96.67 & 94.87 & 73.68 \\
  \midrule
  ACD-U      & 96.46 & 96.28 & 94.64 & \textbf{75.98} \\
  \bottomrule
 \end{tabular}
\end{table}

\begin{table}[t]
  \centering
  \caption{Top-1 and Top-5 accuracies (\%) on the WebVision and ImageNet validation sets.}
  \label{tb:WebVision}
  \begin{tabular}{lcccc}
    \toprule
    Method             & \multicolumn{2}{c}{WebVision} & \multicolumn{2}{c}{ImageNet} \\
    \cmidrule(lr){2-3} \cmidrule(lr){4-5}
                       & Top1 & Top5 & Top1 & Top5 \\
    \midrule
    DivideMix~\cite{Li_2020_DivideMix}  & 77.3      & 91.6      & 75.2      & 90.8      \\
    LongReMix~\cite{cordeiro_2023_LongReMix} & 78.9 & 92.3 & \_   & \_   \\
    RankMatch~\cite{Zhang_2023_Rankmatch}  & 79.9      & \underline{93.6}     & 77.4      & \underline{94.3}      \\
    SV-Learner~\cite{Liang_2024_SV-Learner}  & 78.9      & \underline{93.6}      & 76.9      & 93.9      \\
    NoiseBox+SS-KNN ~\cite{Feng_2024_NoiseBox}   & 81.4      & 93.0      & 77.0      & 91.0      \\
    LSL~\cite{kim_2024_LSL} & 81.4 & 93.0 & 75.8 & 91.8 \\
    Semi-RML++~\cite{li_2025_RML} &  \textbf{83.0}& - & \underline{78.7} & - \\
    CLIPCleaner~\cite{Feng_2024_Clipcleaner} & 81.6      & 93.3      & 77.8      & 92.1      \\
    \midrule
    ACD-U     & \underline{82.8} & \textbf{94.9} & \textbf{81.0} & \textbf{95.5} \\
    \bottomrule
  \end{tabular}
\end{table}

\begin{table}[t]
  \centering
  \caption{Top-1 accuracies (\%) comparison on the Clothing1M test set.}
  \label{tb:Clothing1M}
  \begin{tabular}{lc}
    \toprule
    Method & Top1 (\%) \\
    \midrule
    DivideMix~\cite{Li_2020_DivideMix}    & 74.8      \\
    LongReMix~\cite{cordeiro_2023_LongReMix} & 74.4  \\
    ProMix~\cite{xiao_2022_promix}    & 74.9      \\
    RankMatch~\cite{Zhang_2023_Rankmatch}    & 75.2      \\
    SV-Learner~\cite{Liang_2024_SV-Learner}    & 75.2      \\
    NoiseBox+SS-KNN~\cite{Feng_2024_NoiseBox}   & 74.7      \\
    Semi-RML++~\cite{li_2025_RML} &  75.4\\
    CLIPCleaner~\cite{Feng_2024_Clipcleaner}    & 74.9      \\
    \midrule
    ACD-U       & \textbf{75.5} \\
    \bottomrule
  \end{tabular}
\end{table}

\begin{table}[t]
  \centering
  \caption{Top-1 accuracies (\%) on the Red Mini-ImageNet test set with varying noise rates.}
\label{tb:red_mini}
\begin{tabular}{lcccc}
    \toprule
    Method/ noise ratio & 20\% & 40\% & 60\% & 80\% \\
    \midrule
    DivideMix~\cite{Li_2020_DivideMix} & 50.96 & 46.72 & 43.14 & 34.50 \\
    LongReMix~\cite{cordeiro_2023_LongReMix}& 56.03 & 50.69 & 46.81 & 38.24 \\
    NoiseBox+SS-KNN~\cite{Feng_2024_NoiseBox}  & 60.94 & 59.08 & 50.30 & 44.44 \\
    LSL~\cite{kim_2024_LSL} & 54.68 & 49.80 & 45.46 & 36.78 \\
    CLIPCleaner~\cite{Feng_2024_Clipcleaner} &61.44 & 58.42 & 53.18 & 43.82 \\
    \midrule
    ACD-U  & \textbf{61.52} & \textbf{60.66} & \textbf{57.00} & \textbf{48.94} \\
    \bottomrule
\end{tabular}
\end{table}

\subsubsection{Performance on real-world noise environments}\label{sec:04:02:02}
The evaluation results in real-world noise environments are listed in Tables~\ref{tb:CIFAR-N}---~\ref{tb:red_mini}. On CIFAR-N, our findings mirror the synthetic-noise results. While ProMix shows the highest performance on the simpler CIFAR-10N dataset (with a maximum 1.7\% difference from ACD-U), ACD-U achieves the highest performance on the more challenging CIFAR-100N benchmark, which contains human annotation errors resulting in a 40.20\% noise rate. Under these conditions, our method outperforms LSL by 1.5\% and Semi-RML++ by 2.3\%. This trend aligns with CIFAR-10 artificial noise results, reconfirming that improvement limits are more apparent in simpler tasks.

On the large-scale WebVision dataset, our method demonstrates exceptional generalization. It not only achieves the highest Top-5 accuracy on the WebVision test set but also outperforms Semi-RML++ by 2.3\% in Top-1 accuracy when evaluated on ImageNet. This strong cross-dataset performance indicates that ACD-U learns robust, generalizable features rather than simply filtering noise.

The framework’s robustness is further confirmed on Clothing1M, a million-scale dataset with complex instance-dependent noise, where it achieves the highest performance. While the margin over Semi-RML++ is only 0.1\%, this represents a significant improvement, given the dataset's scale and complexity. On Red Mini-ImageNet, ACD-U's superiority is even more pronounced, as it outperforms all existing methods at all noise rates. For instance, at an 80\% noise rate, our method outperforms NoiseBox+SS-KNN and CLIPCleaner by 4.50\% and 5.12\%, respectively. This robustness is also evident in its resilience to increasing noise; ACD-U's performance degrades by only 12.58\% as noise increases from 20\% to 80\%, a marked improvement over the 16.46\% degradation for DivideMix and 17.62\% for CLIPCleaner.

\noindent \textbf{Summary of answer to RQ1.} Across diverse noise conditions, including high noise rates, complex classification tasks, and real-world label noise, ACD-U achieved performance that is comparable or superior to that of state-of-the-art methods. The results validate the effectiveness of combining asymmetric co-teaching with a novel unlearning mechanism, confirming our approach provides a robust solution for learning in both synthetic and real-world noisy environments.

\subsection{Effectiveness verification of main technologies in the proposed method (RQ2)}\label{sec:04:03}
To answer RQ2, we conducted a series of ablation studies to quantitatively analyze the individual and synergistic effects of our proposed technologies. We report on the individual contributions of unlearning and ACD (Section~\ref{sec:04:03:01}), analyze their impact on performance throughout the training process (Section~\ref{sec:04:03:02}), and evaluate their effect on sample-selection accuracy (Section~\ref{sec:04:03:03}).

\subsubsection{Individual effect analysis of unlearning and ACD}\label{sec:04:03:01}
Table~\ref{tb:Ablation_unlearning_ACD} shows the distinct contributions of each core component on CIFAR-100 at Sym.50\% and Sym.80\%, revealing that unlearning and ACD excel under different conditions. Disabling the unlearning component has a minimal impact at Sym.50\% (a <0.2\% decrease in Last accuracy), but its removal causes a significant performance drop of 1.7\% in Best accuracy and 1.0\% in Last accuracy at Sym.80\%. This indicates that selective forgetting is most critical in high-noise environments. Conversely, disabling the ACD framework  has a significant impact at Sym.50\%, causing a 1.9\% decrease in both Best and Last accuracies, whereas its effect is minimal at Sym.80\%, with a 0.7\% improvement in Last. These results suggest that the two components play complementary roles.

\subsubsection{Analysis of performance transitions during learning}\label{sec:04:03:02}
\begin{figure*}[t]
  \centering
  \subfigure[CIFAR-100 (Sym. 50\%)]{\includegraphics[width=0.45\textwidth]{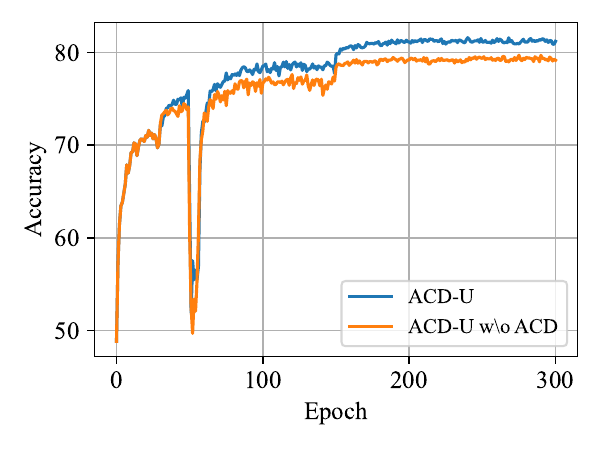}}
  \subfigure[CIFAR-100 (Sym. 80\%)]{\includegraphics[width=0.45\textwidth]{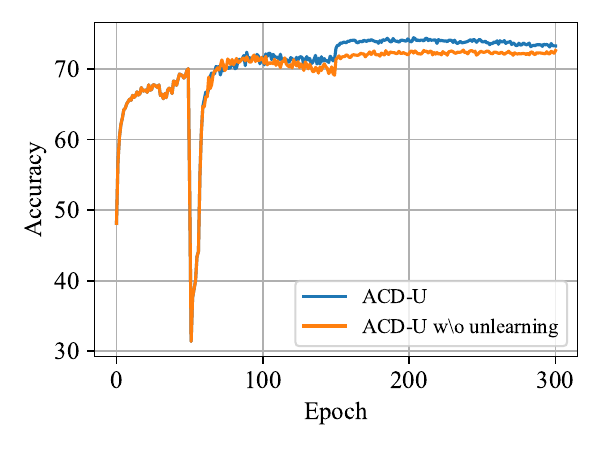}}
 \caption{Accuracy curves on CIFAR-100 with (a) 50\% and (b) 80\% symmetric noise, showing the effects of the ACD and unlearning components, respectively.}
  \label{fig:Ablation_unlearning_ACD}
\end{figure*}

Fig.~\ref{fig:Ablation_unlearning_ACD} visualizes the accuracy curves from our ablation study to illustrate the learning dynamics. The plots specifically show the impact of the ACD component on CIFAR-100 (Sym.50\%) and the unlearning component on CIFAR-100 (Sym.80\%). These settings were selected for visualization as they represent the conditions where Table~\ref{tb:Ablation_unlearning_ACD} demonstrated the most significant performance degradation when each respective component was removed.
The temporary accuracy dip observed around epoch 50 in all runs corresponds to the unfreezing of the ViT encoder in net $V$. At Sym.50\%, the full ACD-U model consistently outperformed the variant without ACD after the warmup period. By training the stable, pretrained net $V$ only on clean samples, ACD effectively suppresses the memorization of noisy labels in early-to-mid training, leading to better overall performance.

A different pattern emerges at Sym.80\%. Between epochs 100 and 150, the model without unlearning begins to suffer from performance degradation, whereas the full ACD-U model maintains stable performance. This demonstrates that as the number of incorrectly memorized samples increases in high-noise settings, the ability to dynamically detect and forget these samples is crucial for preventing overfitting and maintaining performance in the subsequent training stages.

This analysis confirms the complementary nature of our contributions. The ACD framework provides its greatest benefit (approximately 2\% improvement) in moderate-noise environments, such as Sym.50\%, by leveraging the stable predictions of the pretrained ViT on clean samples. By contrast, the unlearning mechanism becomes the key contributor in high-noise environments such as Sym.80\%, providing a 1.7\% improvement by actively correcting memorized errors.

ACD's effect is limited in high-noise environments because, with numerous noisy samples, active correction via unlearning becomes more critical than careful initial selection. Conversely, the impact of unlearning is less pronounced in low-to-medium noise environments owing to fewer incorrectly memorized samples that require correction. By combining both technologies, ACD-U achieves robust and stable performance across a wide range of noise conditions.

\begin{table}[t]
\centering
\caption{Ablation analysis of the unlearning and ACD components on CIFAR-100 (Sym.50\%, 80\%). Values in parentheses denote the change in accuracy compared with the full ACD-U model.}
\label{tb:Ablation_unlearning_ACD}
\begin{tabular}{l l cc}
\toprule
\multicolumn{2}{l}{Method} & \multicolumn{2}{c}{\textbf{CIFAR-100 (Sym.)}} \\
\cmidrule(lr){3-4}
\multicolumn{2}{l}{} & 50\% & 80\% \\
\midrule
\multirow{2}{*}{ACD-U} 
    & Best & \textbf{81.6} & \textbf{74.4} \\
    & Last & \textbf{81.2} & \underline{73.3} \\
\multirow{2}{*}{ACD-U w/o unlearning} 
    & Best & \textbf{81.6}~($\pm0.0$) & 72.7~($-1.7$) \\
    & Last & \underline{81.0}~($-0.2$) & 72.3~($-1.0$) \\
\multirow{2}{*}{ACD-U w/o ACD} 
    & Best & 79.7~($-1.9$) & \textbf{74.4}~($\pm0.0$) \\
    & Last & 79.3~($-1.9$) & \textbf{74.0}~($+0.7$) \\
\bottomrule
\end{tabular}
\end{table}

\begin{figure*}[t]          
  \centering
  \includegraphics[width=0.9\textwidth, height=0.45\textheight, keepaspectratio]
  {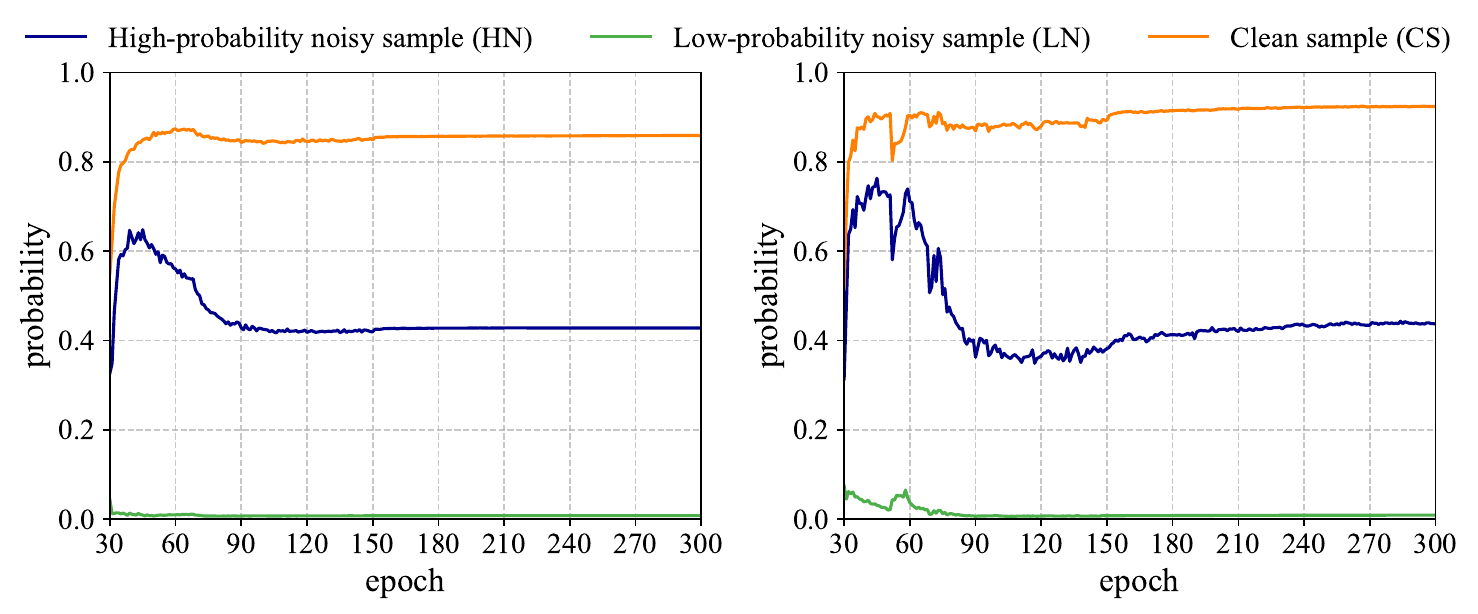}
 \caption{Comparison of how noisy samples misidentified as clean (HN) are treated over time on CIFAR-100 (Sym. 80\%). Left: In DivideMix, initial errors persist. Right: In ACD-U, initial errors are corrected.}
  \label{fig:Ablation_prediction}
\end{figure*}

\begin{table}[t]
\centering
\caption{Number of misclassified samples in each category during early training on CIFAR-100 (Sym.80\%). HN: noisy samples incorrectly judged as clean; LN: noisy samples correctly judged as noisy; CS: clean samples.}
\label{tb:Ablation_prediction}
\renewcommand{\arraystretch}{0.86}
\begin{tabular}{l ccc}
\toprule
Method & HN & LN & CS \\
\midrule
DivideMix & 6444 & 33167 & 10389 \\
ACD-U  & 1127 & 38484 & 10389 \\
\bottomrule
\end{tabular}
\end{table}

\subsubsection{Improvement in sample-selection accuracy}\label{sec:04:03:03}
The ACD framework, utilizing a CLIP-based pretrained ViT, is designed to improve sample-selection accuracy during the critical early stages of training. The effectiveness of this approach is detailed in Table~\ref{tb:Ablation_prediction}, which compares the classification of noisy and clean samples for ACD-U and DivideMix during early training (epochs 30–50) on CIFAR-100 (Sym.80\%). For this analysis, we report on noisy samples incorrectly judged as clean (HN), those correctly judged as noisy (LN), and clean samples (CS). A sample is  defined as incorrectly judged as clean if both networks assign it a clean probability $w_i^{(l)}$ $(l \in \{A, V\})$ of 0.5 or higher at least once during this period. The results show that ACD-U reduces the number of these critical HN samples to approximately one-sixth of those produced by DivideMix, confirming that the pretrained ViT significantly suppresses such misjudgments during the early training stages.

Fig.~\ref{fig:Ablation_prediction} illustrates the long-term impact of these initial selections by tracking the proportion of samples that net $A$ continues to judge as clean ($w_i^{(A)} \geq 0.5$) over time. For DivideMix, the analysis reveals that once a noisy sample is misidentified as clean (HN), it tends to remain so; over 40\% of these initial errors are never corrected, demonstrating that they become permanently memorized as noisy. By contrast, ACD-U demonstrates a dynamic error correction process. While the proportion of HNs judged as clean temporarily exceeds that of DivideMix between epochs 40--60, it subsequently decreases significantly (below 40\%) at epochs 90--150, eventually converging to DivideMix's level. This trajectory indicates that the unlearning mechanism successfully identifies and corrects these initial selection errors, thereby preventing incorrect memorization.

\noindent \textbf{Summary of answer to RQ2.} Our ablation studies revealed that unlearning and ACD make distinct and complementary contributions to performance. Unlearning is most critical in high-noise environments (Sym.80\%), where it provides a 1.7\% performance improvement by correcting incorrectly memorized samples. By contrast, ACD is most impactful at moderate noise rates (Sym.50\%), where it contributes a 1.9\% improvement by suppressing early-stage noise memorization. The integration of a CLIP-pretrained ViT via the ACD framework substantially improves sample-selection accuracy, reducing critical misclassifications to approximately one-sixth of DivideMix's level. This powerful synergy between better initial selection and post-hoc error correction enables ACD-U to maintain robust performance across diverse noise conditions.

\subsection{Design choices and parameter optimization (RQ3)}\label{sec:04:04}
To answer RQ3, we systematically analyzed the effects of our unlearning selection conditions, key hyperparameters, and learning strategies on overall performance. This section reports on the impact of the selection conditions (Section~\ref{sec:04:04:01}), hyperparameter sensitivity (Section~\ref{sec:04:04:02}), and strategic design choices (Section~\ref{sec:04:04:03}).

\subsubsection{Effects of forgetting target sample selection conditions}\label{sec:04:04:01}
Our unlearning sample-selection mechanism integrates three conditions: low-loss (Eq.~\eqref{eq:low_loss_quantile}), loss-drop (Eq.~\eqref{eq:loss_change}), and CLIP-consistent (Eq.~\eqref{eq:CLIP_selection}). Table~\ref{tb:Ablation_unlearning_SS} presents the impact of removing each condition on CIFAR-100 (Sym.80\%).

The results indicate that the CLIP-consistent condition is the most critical component. Excluding it causes the most significant performance drop, with 2.6\% and 3.9\% degradations in Best and Last accuracies, respectively. This highlights the importance of using the pretrained CLIP model's noise-independent predictions to protect genuinely CS from being unlearned. The low-drop condition is also significant; removing it leads to 1.6\% and 1.2\% drops in Best and Last accuracies, respectively, confirming that the change in loss remains a valid basis for selection even in high-noise environments. Finally, excluding the low-loss condition has a mixed effect, slightly improving Last accuracy by 0.2\% but decreasing Best accuracy by 0.5\%, indicating that it primarily contributes to achieving peak performance and stabilizing the learning process.

\begin{table}[t]
\centering
\caption{Ablation analysis of the three conditions used for unlearning sample selection on CIFAR-100 (Sym.80\%). Numbers in parentheses represent differences from ACD-U.}
\label{tb:Ablation_unlearning_SS}
    \begin{tabular}{l l c}
    \toprule
    \multicolumn{2}{l}{Method} & \multicolumn{1}{c}{\textbf{CIFAR-100 (Sym.)}} \\
    \cmidrule(lr){3-3}
    \multicolumn{2}{l}{} & 80\% \\
    \midrule
    \multirow{2}{*}{ACD-U}
        & Best &  \textbf{74.4} \\
        & Last &     73.3 \\
    \midrule
    \multirow{2}{*}{ACD-U w/o Low-loss}
        & Best &  73.9 (-0.5) \\
        & Last &  \textbf{73.5} (+0.2) \\
    \midrule
    \multirow{2}{*}{ACD-U w/o Loss-drop}
        & Best &  72.8 (-1.6) \\
        & Last &    72.1 (-1.2) \\
    \midrule
    \multirow{2}{*}{ACD-U w/o CLIP-consistent}
        & Best & 71.8 (-2.6) \\
        & Last & 69.4 (-3.9) \\
    \bottomrule
    \end{tabular}
\end{table}

\subsubsection{Sensitivity analysis of hyperparameters}\label{sec:04:04:02}
\begin{figure*}[t]          
  \centering
  \includegraphics[width=0.9\textwidth, height=0.4\textheight, keepaspectratio]{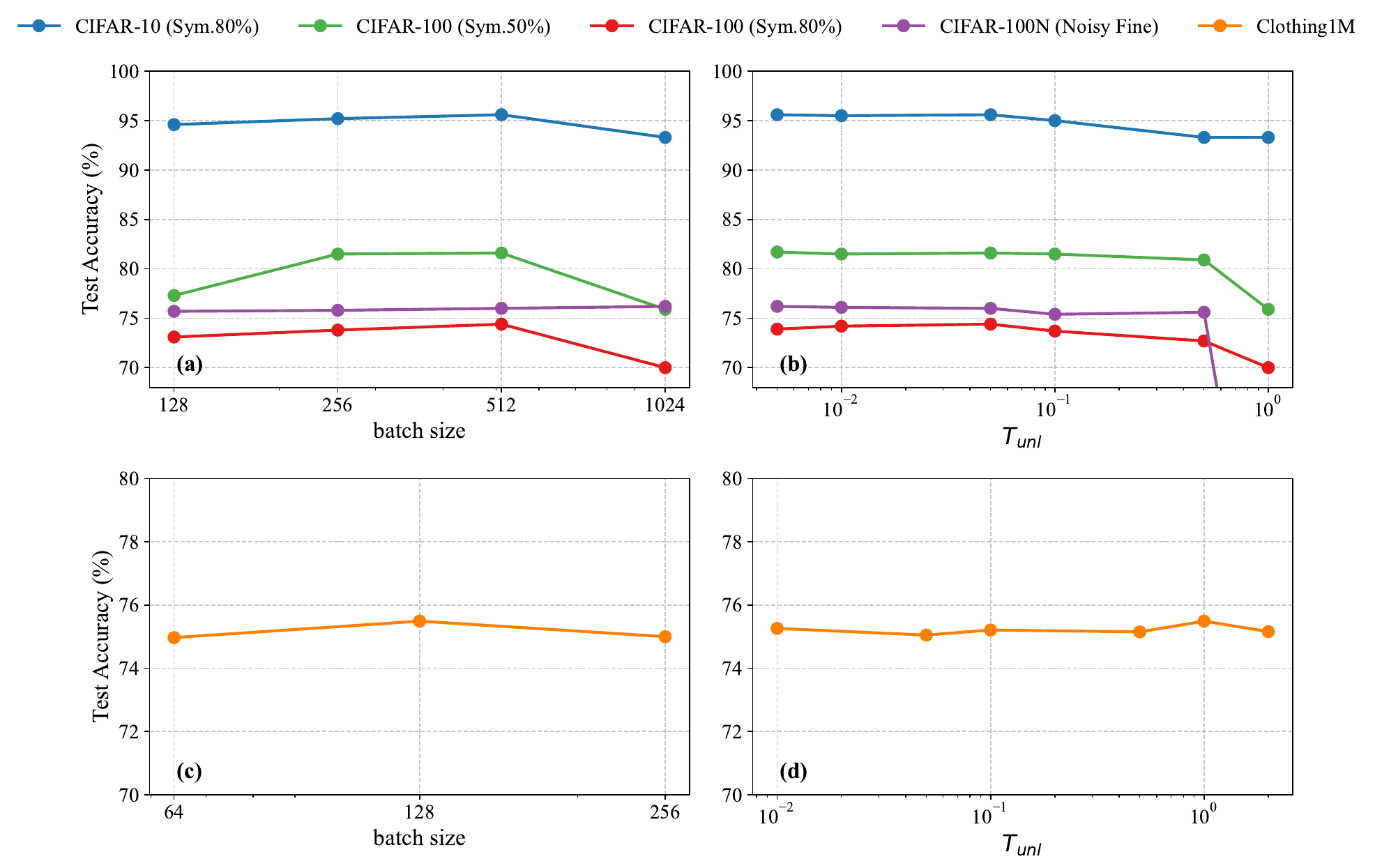}
  \caption{Sensitivity analysis for the unlearning batch size and forgetting intensity parameter $T_{unl}$. Panels (a) and (b) show the results for the CIFAR datasets, whereas (c) and (d) show those for Clothing1M. For the CIFAR-10/-100 datasets, the “Best” accuracy is shown, whereas for CIFAR-100N and Clothing1M, the reported accuracies are based on the same evaluation criteria as those detailed in Tables~\ref{tb:CIFAR-N} and~\ref{tb:Clothing1M}.}
  \label{fig:Ablation_Unlearning_parameter}
\end{figure*}

Fig.~\ref{fig:Ablation_Unlearning_parameter} shows the effects of the unlearning batch size and forgetting intensity parameter, $T_{unl}$. The optimal batch size is dataset-dependent: 512 is best for CIFAR-10/100, whereas 128 is optimal for Clothing1M. The smaller batch size for Clothing1M is likely necessary for the finer control required by its large images ($224\times224$) and complex instance-dependent noise. Performance degrades at the extremes, illustrating a key trade-off where small batch sizes risk excessive forgetting, while large ones may cause insufficient forgetting. For instance, a batch size of 128 results in poor performance on the CIFAR datasets, whereas a size of 1024 leads to frequent early stopping. The forgetting intensity, controlled by the $T_{unl}$ parameter, is also highly sensitive. Performance degrades significantly when $T_{unl}$ is set to 0.1 or higher, with the training process collapsing at 0.5 or 1.0. By contrast, the performance degradation is only moderate when reducing $T_{unl}$ from 0.05 to 0.01. This asymmetry indicates that the unlearning process is more robust to conservative forgetting than to aggressive forgetting. However, Clothing1M is a notable exception, where a considerably higher value of $T_{unl}=1.0$ is optimal. This suggests that its complex, instance-dependent noise requires a stronger corrective signal during the unlearning phase.

\begin{figure*}[t]
  \centering
  \includegraphics[width=0.9\textwidth, height=0.4\textheight, keepaspectratio]{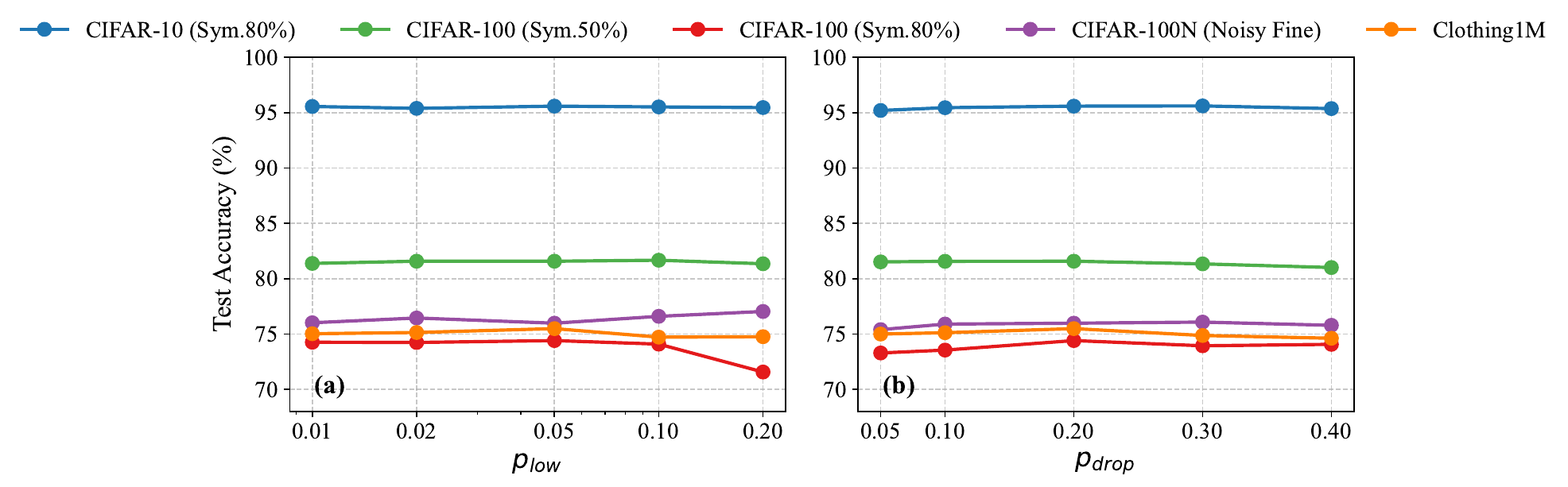}
  \caption{Sensitivity analysis for the unlearning sample-selection hyperparameters (left column: low-loss sample proportion ($p_{low}$), right column: loss-drop sample proportion ($p_{drop}$)). For the CIFAR-10/-100 datasets, the ``Best'' accuracy is shown, whereas for CIFAR-100N and Clothing1M, the reported accuracies are based on the same evaluation criteria as those detailed in Tables~\ref{tb:CIFAR-N} and~\ref{tb:Clothing1M}.}

  \label{fig:Ablation_Unlearning_parameter_unlearningSS}
\end{figure*}

Fig.~\ref{fig:Ablation_Unlearning_parameter_unlearningSS} analyzes the sensitivity to the selection proportion hyperparameters. The model is more sensitive to low-loss sample proportion ($p_{low}$), particularly to increases. On CIFAR-100 (Sym.80\%), increasing $p_{low}$ from 0.05 to 0.2 causes a significant performance degradation. However, the framework is more robust to decreases, showing only moderate degradation when the value is reduced to 0.01.By contrast, the model is relatively robust to the choice of the low-drop sample proportion ($p_{drop}$), although setting the value too low (e.g.,$p_{drop}=0.05$) degrades performance because it cannot properly select the overfitted noisy samples during training.

\subsubsection{Design choices for learning strategies}\label{sec:04:04:03}
We analyze two key strategic design choices: label usage during warmup and the fine-tuning of the ViT encoder. As shown in Table~\ref{tb:ablation_labels}, the warmup strategy for net $V$ has a significant impact. Using self-supervised learning without labels (our default ACD-U approach) yields a 3\% performance improvement on CIFAR-100 (Sym.90\%) than using noisy labels during warmup (ACD-U w/ ULdW). This indicates that supervised warmup causes the model to overfit to noise patterns early, making subsequent correction more difficult. A self-supervised learning approach allows the model to learn  essential, noise-free features first and prevents overfitting.

\begin{table*}[t]
  \centering
  \caption{Ablation analysis comparing two warmup strategies for net $V$ on CIFAR-10/100: self-supervised learning without labels (ACD-U) versus supervised learning with the original noisy labels (ACD-U w/ ULdW).}
  \label{tb:ablation_labels}
  \newcolumntype{C}{>{\centering\arraybackslash}X}
  \begin{tabularx}{\textwidth}{l l *{9}{C}}
    \toprule
    \multicolumn{2}{l}{Dataset} 
      & \multicolumn{5}{c}{\textbf{CIFAR-10}} 
      & \multicolumn{4}{c}{\textbf{CIFAR-100}} \\
    \cmidrule(lr){3-7} \cmidrule(lr){8-11}
    \multicolumn{2}{l}{Noise type} 
      & \multicolumn{4}{c}{Sym.} & Asym. 
      & \multicolumn{4}{c}{Sym.} \\
    \multicolumn{2}{l}{Method/ Noise rate} 
      & 20\% & 50\% & 80\% & 90\% & 40\% 
      & 20\% & 50\% & 80\% & 90\% \\
    \midrule
    \multirow{2}{*}{ACD-U}  
      & best & 97.2 & \textbf{96.8} & \textbf{95.6} & 93.9 & 95.7 
              & \textbf{83.3} & \textbf{81.6} & \textbf{74.4} & \textbf{66.5} \\
      & last & \textbf{97.0} & \textbf{96.6} & \textbf{95.4} & 93.3 & 94.4 
              & \textbf{83.0} & \textbf{81.2} & 73.3 & \textbf{63.4} \\
    \midrule
    \multirow{2}{*}{ACD-U w/ ULdW} 
      & best & \textbf{97.3} & 96.7 & \textbf{95.6} & \textbf{94.0} &  \textbf{95.9}
              & \textbf{83.3} & 81.5 & 74.0 & 61.6 \\
      & last & \textbf{97.0} & \textbf{96.6} & 95.3 & \textbf{93.5} & \textbf{94.9} 
              & 82.9 & 81.0 & \textbf{73.7} & 60.9 \\
    \bottomrule
  \end{tabularx}
\end{table*}

Table~\ref{tb:ablation_encoder} shows that the decision to fine-tune the ViT encoder involves a critical trade-off that depends on the noise conditions. While gradually fine-tuning the encoder (our default ACD-U approach, which starts gradient computation at epoch 50) is beneficial in low-noise environments because it leverages the ViT’s full expressive power, this process can destabilize the model’s predictions in high-noise environments, leading to incorrect sample selection.  Consequently, keeping the encoder frozen (ACD-U w/ stop-grad) is the superior strategy under  high noise. Therefore, we suggest avoiding fine-tuning for high-noise and complex datasets, while applying it in low-to-medium noise environments. This is reflected in our experiments: we kept the encoder frozen for the challenging Red Mini-ImageNet, whereas for CIFAR-10/100, fine-tuning from epoch 50 provided a clear benefit at lower noise rates.

\begin{table}[t]
  \centering
  \caption{Ablation analysis comparing the effect of fine-tuning versus freezing the ViT encoder for net $V$ on CIFAR-100}
  \label{tb:ablation_encoder}
  \newcolumntype{C}{>{\centering\arraybackslash}X}
  \begin{tabularx}{\columnwidth}{l l *{4}{C}}
    \toprule
    \multicolumn{2}{l}{Dataset}
      & \multicolumn{4}{c}{\textbf{CIFAR-100}} \\
    \cmidrule(lr){3-6}
    \multicolumn{2}{l}{Noise type}
      & \multicolumn{4}{c}{Sym.} \\
    \multicolumn{2}{l}{Method/ Noise rate}
      & 20\% & 50\% & 80\% & 90\% \\
    \midrule
    \multirow{2}{*}{ACD-U}
      & best& \textbf{83.3} & \textbf{81.6} & 74.4 & 66.5 \\
      & last & \textbf{83.0} & \textbf{81.2} & 73.3 & 63.4 \\
    \midrule
    \multirow{2}{*}{ACD-U  w/ stop-grad}
      & best & 81.0 & 80.5 & \textbf{75.1} & \textbf{70.9} \\
      & last & 80.5 & 79.7 & \textbf{74.6} & \textbf{70.5} \\
    \bottomrule
  \end{tabularx}
\end{table}

\noindent \textbf{Summary of answer to RQ3.} Our systematic analysis revealed that optimal design guidelines emerge from the experiments: (1) The CLIP-Consistent condition is essential for performance, demonstrating maximum effectiveness when integrated with other selection conditions; (2) the unlearning batch size requires adjustment based on dataset complexity (e.g.,  512 for CIFAR and 128 for Clothing1M); (3) a baseline $T_{unl}$ of 0.05 is optimal for most datasets, whereas larger values are required for complex noise patterns; (4) the ViT encoder should remain frozen at high noise rates, whereas gradual fine-tuning is beneficial in low-to-medium noise scenarios. These interrelated design choices necessitate comprehensive adjustments according to dataset characteristics to maximize performance.

\section{Conclusions}\label{sec:05}
This paper presented ACD-U, a novel framework that systematically integrates machine unlearning and asymmetric co-teaching to address two fundamental challenges in LNL: the difficulty of correcting memorized errors post-hoc and suboptimal utilization of pretrained models with different learning characteristics. ACD-U is built on two primary technical innovations. First, it introduces a practical machine unlearning pipeline for the LNL domain, with a framework that dynamically identifies and forgets incorrectly memorized noisy samples by analyzing model-loss changes and consistency with CLIP's zero-shot predictions. Second, it employs an asymmetric co-teaching architecture that pairs a CLIP-based pretrained ViT with a conventional CNN, a design that provides complementary mitigation of confirmation bias.

Our extensive experiments demonstrated the effectiveness of ACD-U. On synthetic benchmarks, it achieved a 35\% relative improvement over DivideMix on CIFAR-100 at Sym.90\%. Moreover, it delivered state-of-the-art performance on real-world noise datasets, outperforming LSL by 1.5\% and Semi-RML++ by 2.3\% on CIFAR-100N, and surpassing all existing methods on Red Mini-ImageNet across all noise rates. Furthermore, on WebVision, it achieved the highest Top-5 accuracy and outperformed Semi-RML++ by 2.3\% in a cross-dataset evaluation on ImageNet, while also recording the top performance on Clothing1M, where it outperformed Semi-RML++ by 0.1\%. Ablation analysis quantitatively confirmed that the unlearning and ACD components play complementary roles. The unlearning mechanism contributed a 1.7\% performance improvement in high-noise environments (Sym.80\%) by  effectively correcting incorrectly memorized samples. Concurrently, it achieved a 2\% performance improvement in low-to-medium noise rate environments (Sym.50\%) by successfully suppressing noise memorization during the early training stages. Finally, the introduction of a pretrained ViT in our framework led to a significant improvement in sample-selection accuracy. In our analysis, ACD-U reduced the number of misjudged samples to approximately one-sixth of those made by DivideMix, demonstrating significant improvement in selection accuracy through the introduction of pretrained ViT.

Our analysis also identified several limitations. First, on simple tasks such as CIFAR-10, the performance gains over ProMix remained modest (0.5--0.6\%), highlighting the challenge of achieving significant improvements in low-complexity environments. Second, because the framework relies on a CLIP-based ViT, its performance may be limited in specialized domains not included in CLIP's pretraining data. Finally, some key hyperparameters, particularly the unlearning intensity ($T_{unl}$), have narrow optimal ranges that require careful tuning for new datasets; for instance, performance degrades significantly when $T_{unl}$ is set to 0.1 or higher.

Based on our findings, we identify several avenues for future work. First, to improve the framework’s generalizability, exploring ACD with other architectural combinations beyond the current ResNet and ViT pairing is crucial. Second, the unlearning sample-selection process could be enhanced by incorporating historical information, for example, through the use of memory banks. Finally, a comparative study of alternative unlearning mechanisms beyond the  KL divergence-based approach could lead to more efficient and effective forgetting strategies.

\bibliographystyle{elsarticle-num} 
\bibliography{references} 

\appendix

\section{GMM for sample selection}
Our sample-selection mechanism employs a two-component GMM, following the approach of DivideMix~\cite{Li_2020_DivideMix}, to distinguish between clean and noisy samples based on their loss distributions. This method leverages the empirical observation that CS typically exhibits lower losses than noisy samples during training.

\subsection{GMM algorithm}
Given per-sample losses $\{\ell_i\}_{i=1}^N$ from a network, we model the loss distribution as a mixture of two Gaussian components that represent the clean and noisy subsets of the data:
\begin{equation}
p(\ell_i) = \pi_c \mathcal{N}(\ell_i; \mu_c, \sigma_c^2) + \pi_n \mathcal{N}(\ell_i; \mu_n, \sigma_n^2),
\end{equation}
where $\pi_c$ and $\pi_n = 1 - \pi_c$ denote the mixing coefficients, and  $(\mu_c, \sigma_c^2)$ and $(\mu_n, \sigma_n^2)$ represent the mean and variance parameters for the clean and noisy components, respectively.

The GMM parameters are estimated using the expectation-maximization algorithm, as detailed in Algorithm~\ref{alg:gmm}. This algorithm iteratively updates the posterior probabilities (E-step) and component parameters (M-step) until convergence is achieved. Based on the principle that CS has lower losses, the Gaussian component with the smaller mean is designated as the clean component.

\begin{algorithm}[h]
\SetAlgoLined
\caption{GMM Parameter Estimation}
\label{alg:gmm}
\KwIn{per-sample losses $\{\ell_i\}_{i=1}^N$, maximum iterations $T_{max}$, convergence threshold $\epsilon$}
\KwOut{clean probabilities $\{w_i\}_{i=1}^N$}

Initialize parameters $\theta^{(0)} = \{\pi_1^{(0)}, \pi_2^{(0)}, \mu_1^{(0)}, \mu_2^{(0)}, \sigma_1^{(0)}, \sigma_2^{(0)}\}$\;

\For{$t = 1$ \KwTo $T_{max}$}{
   \tcp{E-step: Compute posterior probabilities}
   \For{$i = 1$ \KwTo $N$}{
       $\gamma_{i1}^{(t)} = \frac{\pi_1^{(t-1)} \mathcal{N}(\ell_i; \mu_1^{(t-1)}, (\sigma_1^{(t-1)})^2)}{\pi_1^{(t-1)} \mathcal{N}(\ell_i; \mu_1^{(t-1)}, (\sigma_1^{(t-1)})^2) + \pi_2^{(t-1)} \mathcal{N}(\ell_i; \mu_2^{(t-1)}, (\sigma_2^{(t-1)})^2)}$\;
       $\gamma_{i2}^{(t)} = 1 - \gamma_{i1}^{(t)}$\;
   }
   
   \tcp{M-step: Update parameters}
   $N_1^{(t)} = \sum_{i=1}^N \gamma_{i1}^{(t)}$, $N_2^{(t)} = \sum_{i=1}^N \gamma_{i2}^{(t)}$\;
   
   $\pi_1^{(t)} = \frac{N_1^{(t)}}{N}$, $\pi_2^{(t)} = \frac{N_2^{(t)}}{N}$\;
   
   $\mu_1^{(t)} = \frac{\sum_{i=1}^N \gamma_{i1}^{(t)} \ell_i}{N_1^{(t)}}$, $\mu_2^{(t)} = \frac{\sum_{i=1}^N \gamma_{i2}^{(t)} \ell_i}{N_2^{(t)}}$\;
   
   $(\sigma_1^{(t)})^2 = \frac{\sum_{i=1}^N \gamma_{i1}^{(t)} (\ell_i - \mu_1^{(t)})^2}{N_1^{(t)}}$\;
   $(\sigma_2^{(t)})^2 = \frac{\sum_{i=1}^N \gamma_{i2}^{(t)} (\ell_i - \mu_2^{(t)})^2}{N_2^{(t)}}$\;
   
   \If{$||\theta^{(t)} - \theta^{(t-1)}||_2 < \epsilon$}{
       break\;
   }
}

\tcp{Assign clean probabilities based on the component with the smaller mean}
\If{$\mu_1^{(t)} < \mu_2^{(t)}$}{
   $w_i = \gamma_{i1}^{(t)}$ for all $i$\;
}
\Else{
   $w_i = \gamma_{i2}^{(t)}$ for all $i$\;
}

\Return $\{w_i\}_{i=1}^N$\;
\end{algorithm}

\subsection{GMM-based sample selection in ACD-U}
Within the ACD-U framework, the GMM is applied independently to each network's loss distribution to obtain clean probabilities for enabling cross-network sample partitioning. The process is as follows:
\begin{enumerate}
\item For each network $l \in \{A, V\}$, we compute the cross-entropy loss $\{\ell_i^{(l)} = -\log p_{\theta_l}(y_i|\mathbf{x}_i)\}$ for every sample in the current training set $D_t$.

\item We then apply the GMM estimation from Algorithm~\ref{alg:gmm} to this loss distribution to obtain a set of posterior probabilities, $\{w_i^{(l)}\}$, which represent the likelihood that each sample is clean.

\item Using a threshold $\tau_w$, we partition the samples via cross-network selection (see Fig.~\ref{fig:architecture} for an overview):
    \begin{itemize}
        \item[--] Network $A$ is trained on samples that are partitioned based on the assessments from network $V$.
        \item[--] Network $V$ is trained on samples that are partitioned based on the assessments from network $A$.
    \end{itemize}
    
\item Finally, we execute the asymmetric training step: network $A$ performs SSL using both its labeled ($w_i^{(V)} \geq \tau_w$) and unlabeled ($w_i^{(V)} < \tau_w$) sets, whereas network $V$ performs supervised learning using only its labeled set($w_i^{(A)} \geq \tau_w$). 
\end{enumerate}
This cross-network selection mechanism is crucial for preventing each model from reinforcing its own classification errors, thereby mitigating the confirmation bias that is inherent in single-model or symmetrically trained approaches.

\end{document}